\documentclass[10pt,twocolumn,letterpaper]{article}

\usepackage{nnnn}
\usepackage{times}
\usepackage{epsfig}
\usepackage{graphicx}
\usepackage{amsmath}
\usepackage{amssymb}
\usepackage{multirow}
\usepackage{adjustbox}
\usepackage{makecell}
\usepackage{subfigure}

\usepackage[pagebackref=true,breaklinks=true,letterpaper=true,colorlinks,bookmarks=false]{hyperref}

 \nnnnfinalcopy 

\def\nnnnPaperID{****}

\ifnnnnfinal\fi

\begin{document}

\title{LightSAL: Lightweight Sign Agnostic Learning for Implicit Surface Representation}

\author{Abol Basher\\
University of Vaasa\\
Vaasa, Finland\\
{\tt\small abol.basher@uwasa.fi}
\and
Muhammad Sarmad\\
NTNU\\
Trondheim, Norway\\
{\tt\small muhammad.sarmad@ntnu.no}

\and
Jani Boutellier\\
University of Vaasa\\
Vaasa, Finland\\
{\tt\small jani.boutellier@uwasa.fi}
}

\maketitle
\ifnnnnfinal\thispagestyle{empty}\fi


\begin{abstract}
Recently, several works have addressed modeling of 3D shapes using deep neural networks to learn implicit surface representations. Up to now, the majority of works have concentrated on reconstruction quality, paying little or no attention to model size or training time.
This work proposes LightSAL, a novel deep convolutional architecture for learning 3D shapes; the proposed work concentrates on efficiency both in network training time and resulting model size. We build on the recent concept of Sign Agnostic Learning for training the proposed network, relying on signed distance fields, with unsigned distance as ground truth. In the experimental section of the paper, we demonstrate that the proposed architecture outperforms previous work in model size and number of required training iterations, while achieving equivalent accuracy. Experiments are based on the D-Faust dataset that contains 41k 3D scans of human shapes. The proposed model has been implemented in PyTorch.  
\end{abstract}

\section{Introduction}
Representation of 3D geometry has numerous applications in computer vision, robotics, and computer graphics. Traditional data types for representing 3D geometry include point clouds, voxels, and meshes, each of these formats having their shortcomings: point cloud data is disconnected, the memory occupancy of voxels grows exponentially with resolution, and meshes have issues related to continuity.

Recently popularized \textit{implicit representations} of 3D shapes attempt to address the shortcomings of the current data types. With the introduction of deep learning for implicit shape representation, this research area has progressed rapidly in the last few years \cite{Occupancy_Networks,park2019deepsdf,atzmon2021sald,Peng2020ECCV,chibane2020ndf}. However, learning shapes and surfaces from unstructured and incomplete 3D raw point clouds, triangle soups, or non-manifold meshes is a complex problem: solutions need to encode complex geometry while being cost-effective in terms of computation and memory \cite{Peng2020ECCV}. 

\begin{figure}
   \centering
\begin{tabular}{cc}
\includegraphics[width=3.0cm]{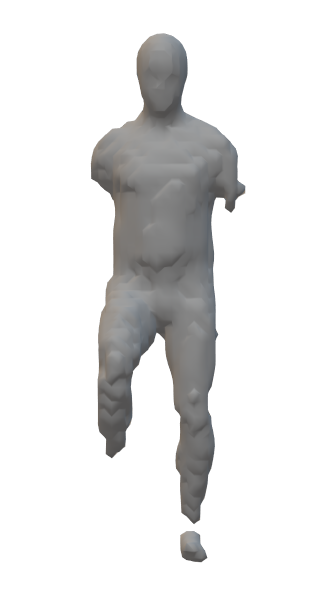}&
\includegraphics[width=3.0cm]{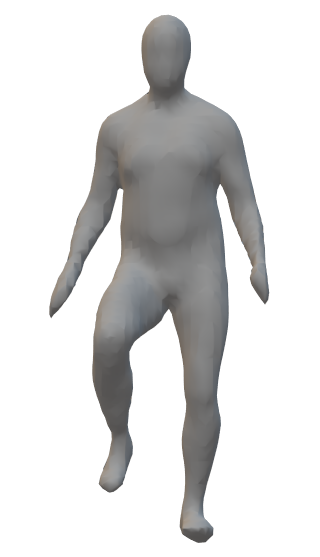}\\
\end{tabular}
    \caption{Point-cloud reconstruction results for 500 training epochs. Left: SAL architecture \cite{Atzmon_2020_CVPR}, Right: proposed LightSAL architecture. }
    \label{fig:teaser}
\end{figure}

Various methods can represent the surface of a shape. A neural network can store information such as occupancy, signed distance, or unsigned distance  \cite{chibane2020implicit,Occupancy_Networks,park2019deepsdf}. In case of occupancy and signed distance, we can reveal the final shape through post-processing, e.g., using the Marching Cubes algorithm \cite{lorensen1987marching}. However, the ground-truth generation for this representation learning requires closed surfaces. Unsigned distance field ground-truth generation does not require closed surfaces. However, the post-processing to obtain the final surface e.g. in the form of a mesh or a point cloud is not straightforward. 

In the implicit representation approach \cite{Atzmon_2020_CVPR,atzmon2021sald,Peng2020ECCV,chibane2020ndf,Occupancy_Networks}, 3D shapes/surfaces are expressed as zero level sets $S$ (Equation~\ref{zero-label-set}) \begin{align}
 S = \{x \in \mathbb{R}^3 | f(x;w)=0\}
\label{zero-label-set}
\end{align}
of a neural implicit function, $f:\mathbb{R}^3\longrightarrow\mathbb{R}$, where $x$ and $w$ represent input data samples and neural network weights, respectively. The surface representations are learned from a signed distance function \cite{park2019deepsdf}, occupancy function \cite{Peng2020ECCV,chibane2020implicit,Occupancy_Networks}, or directly from a raw point cloud or  triangle soup \cite{Atzmon_2020_CVPR,atzmon2021sald}. Most of the implicit representation learning methods rely on data sampled from the ground-truth implicit representation, i.e., signed distance function or occupancy function \cite{Peng2020ECCV,Occupancy_Networks,chibane2020implicit}, which introduces an additional data pre-processing step \cite{berger2017survey}. 

Recently, two novel approaches for \textit{sign agnostic learning} of 3D shapes directly from raw point-clouds or triangle soups have been proposed: SAL~\cite{Atzmon_2020_CVPR} and SALD~\cite{atzmon2021sald}. These approaches can directly learn 3D surface representations without requiring the training data to be sampled from the ground truth implicit representation, in contrast to signed distance functions and occupancy functions. However, the SAL~\cite{Atzmon_2020_CVPR} and SALD~\cite{atzmon2021sald} works rely on a fully-connected neural network architecture that contains 4.2M trainable parameters, and requires as much as 2000 training epochs \cite{Atzmon_2020_CVPR} to produce high-quality results.

This work proposes a lightweight neural network architecture \textit{LightSAL} for sign agnostic learning of implicit 3D representations, which achieves the same visual quality than the original SAL architecture \cite{Atzmon_2020_CVPR} with just 500 training epochs. Figure~\ref{fig:teaser} illustrates the point-cloud reconstruction quality difference in the case that both the original SAL \cite{Atzmon_2020_CVPR} architecture and the proposed LightSAL architecture have been trained for 500 epochs. \textit{Moreover, in Section~\ref{ssec:unseen} we show cases where the original SAL model starts to reconstruct the input point-cloud into a significantly different human shape than what the ground truth indicates; LightSAL, in contrast, has not been observed to suffer from this type of unwanted behavior.}

Section~\ref{sec:experiments} of the paper shows experimental results that cover cases of 1) learning shape space from raw scans, 2) generating unseen human shapes, and 3) generating unseen human poses. The results show that LightSAL
\begin{itemize}
    \itemsep0em
    \item Outperforms the baseline in generation quality for unseen shapes and poses, while having
    \item 75\% smaller model size, and requires
    \item 75\% less training epochs for equivalent reconstruction quality. 
\end{itemize}
In a more general sense, we see that our work brings to attention that by careful neural architecture design, the efficiency of implicit representation learning can be greatly improved, which is of essence in promoting their use to practical applications. Some years ago, MobileNets \cite{howard2017mobilenets} similarly proposed lightweight architectures for 2D image recognition, opening up new application areas in mobile vision.

\section{Related work}

Voxel-based data types \cite{choy20163d,ji2017surfacenet,jimenez2016unsupervised,stutz2018learning} non-parametrically represent volumes as a 3D grid, and are widely used for representing 3D shapes and scenes. They are probably the most intuitive extension from 2D raster images into the 3D domain. Voxels also naturally provide a 3D extension for learning paradigms that have initially been applied in 2D, such as convolutional neural networks.
Consequently, voxel-based representations have been used for long in learning based 3D shape/scene reconstruction \cite{varol2018bodynet,zheng2019deephuman,gilbert2018volumetric}. The most natural use case of voxel-based representation learning is \textit{occupancy values} that denote whether a certain voxel is occupied within the shape of interest or not. 
However, due to the cubically growing memory footprint of voxel-based representations, they are practically limited to a certain level of resolutions \cite{tulsiani2017multi,wu20153d,liao2018deep}. Several approaches \cite{dai2017shape,hane2017hierarchical,maturana2015voxnet,riegler2017octnetfusion,tatarchenko2017octree}, such as operating on multiple scales or by the use of octrees, have alleviated the issue with memory scalability, to some extent.   

Another popular data type for 3D representations are point clouds \cite{prokudin2019efficient,fan2017point,yang2019pointflow,lin2018learning,qi2017pointnet,qi2017pointnet++,park2019deepsdf} that have the advantage of being the inherent output format of, e.g., LIDAR-based 3D scanners. Even though point clouds scale better to describe large areas/volumes than voxels, representing many fine details still implies increased memory footprint. Moreover, point clouds are not well-suited for generating watertight surface descriptions \cite{park2019deepsdf, Peng2020ECCV} as point clouds do not contain any connectivity information. 

Finally, 3D meshes offer a more informative data type that also bears information on connectivity between 3D points. Meshes have been used for classification and segmentation \cite{bronstein2017geometric,guo20153d,wang20183d}, and more recently as the output representation for 3D surface reconstruction. For example, by deforming a template, the mesh-based methods can infer shape. However, it makes the mesh-based methods into a single topological representation \cite{lin2019photometric,wang2018pixel2mesh,kanazawa2018learning,ranjan2018generating,pons2015dyna}. In contrast, there are some neural network-based methods \cite{gkioxari2019mesh,groueix2018papier,liao2018deep} that can predict vertices and faces directly from meshes, but they often lack surface continuity and sometimes result in self-intersecting mesh faces. 

Implicit representations of 3D shapes and surfaces is a quickly evolving field of research. Implicit representations are based on zero level-sets (Equation ~\ref{zero-label-set}) of a function, whereas automation of implicit representation construction can be achieved through implicit representation learning. Implicit representation-based methods for learning 3D shape/surface are mainly based on two approaches: (a) binary occupancy grids \cite{Occupancy_Networks,Peng2020ECCV,saito2019pifu,genova2019deep,chibane2020implicit,deng2019nasa}, and (b) distance functions and fields \cite{park2019deepsdf,Atzmon_2020_CVPR,chen2019learning,michalkiewicz2019deep,jiang2020local,chibane2020ndf,atzmon2021sald}. These methods learn to predict either occupancy probabilities or signed/unsigned distances with regard to an implicit surface, for given 3D inputs. A major advantage of implicit models is that they can represent shapes and scenes in a continuous fashion and naturally deal with various complex topologies. However, one significant drawback related to most of these methods is that they require naturally or artificially closed shapes to operate. In addition to shape or surface representation, implicit representation-based methods have also been used for encoding texture information \cite{oechsle2019texture} and 4D reconstruction \cite{niemeyer2019occupancy}. 

Fully connected layers with ReLU \cite{nair2010rectified} activation functions are the default architecture for most previous works on implicit representations. However, it has been shown that careful addition of skip connections can improve the convergence speed and lead to preservation of sharp details \cite{Occupancy_Networks}. NeRF \cite{nerf} showed that the fully connected layer in an implicit neural network benefits from an additional positional encoding layer at the input. Similarly, SIREN \cite{siren} is a significant step in the direction of intelligent implicit neural architecture design. They suggest using the sine activation function with fully connected layers to learn high-frequency details. However, both SIREN and NeRF do not generalize and suffer in performance when sharing the implicit function space \cite{liif}.  To date, the linear layers with ReLU remain the most successful architecture for the encoder-decoder style implicit networks \cite{liif}. Therefore, we also build in this direction.

This work operates in the context of sign agnostic learning \cite{Atzmon_2020_CVPR} that can produce signed implicit representations of 3D surfaces. Our contribution features a neural network architecture that is based on 1D convolutions\footnote{1D convolution (with kernel size = 1) is also considered as a shared multi-layer perceptron (MLP) by many researchers. However, we use the concept of 1D convolution interchangeably with shared MLP in this research work.}, and provides faster convergence at training time and a more compact trained model while preserving the accuracy of the original SAL neural network \cite{Atzmon_2020_CVPR}.

\section{Proposed neural architecture}
In the following, we present our lightweight 1D convolutional neural architecture, LightSAL, for implicit 3D representation, which is directly trainable on raw scans: point clouds or triangle soups. 
For a given raw input, $X\subset \mathbb{R}^3$, we want to optimize the weights $w\in \mathbb{R}^m$ of a shared multi-layer perceptron (MLP) $f(x;w)$, such that its zero level-set approximates, $X$ to the surface, where $f:\mathbb{R}^3 \times \mathbb{R}^m \longrightarrow \mathbb{R}$ is a shared MLP. For example, if the data $X$ holds in a plane, it is assumed that there is a critical weight $w^{*}$ which will reconstruct that plane as a zero level set $f(x;w^{*})$, because surfaces have an approximate tangent plane nearly everywhere \cite{do2016differential}.

In the following, the proposed 1D convolutional layer-based encoder (Subsection~\ref{encoder}) and decoder (Subsection~\ref{Decoder}) architectures are explained in detail along with information related to training and inference  (Subsection~\ref{ssec:training}).
Our encoder architecture is inspired by the fully-connected variant of PointNet \cite{Occupancy_Networks}. On the other hand, the decoder implementation is based on the DeepSDF decoder \cite{park2019deepsdf} variant presented in \cite{Atzmon_2020_CVPR}. However, for both cases, we have modified the original implementations substantially to achieve an expressive and compact architecture by using 1D convolutional layers with kernel size 1.  


\begin{figure}[t]
\begin{center}
\centering\includegraphics[width=\columnwidth]{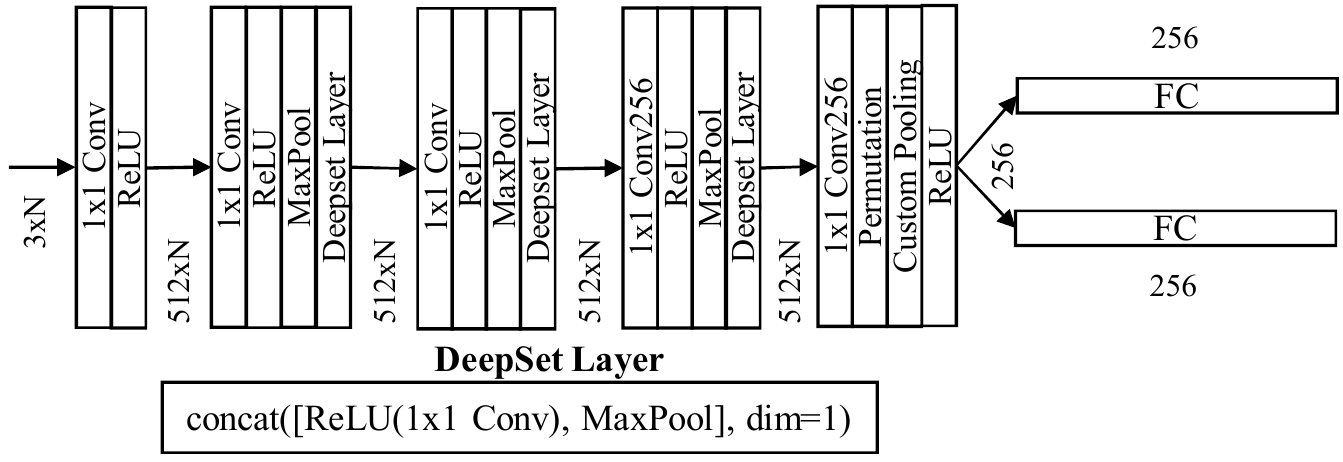}
\caption{The proposed LightSAL encoder network architecture. $N$ at the input stands for number of points in the point cloud.}
\label{fig:encoder}
\end{center}
\end{figure}

\subsection{Encoder}
\label{encoder}

The encoder structure of LightSAL is based on the PointNet \cite{qi2017pointnet} architecture, used in Occupancy Networks~\cite{Occupancy_Networks}. The encoder receives an input point cloud uniformly sampled from raw scans, where each raw scan, $X_{i}\subset\mathbb{R}^3$, and outputs two 256 dimensional vectors, $\mu \in \mathbb{R}^{256}$ and $\eta \in \mathbb{R}^{256}$, that are used to parameterize multivariate Gaussian $\mathcal{N}(\mu,diag\ exp\ \eta)$ for sampling a latent code, $ z\in\mathbb{R}^{256}$. However, in the proposed work, the fully-connected layers used in the baseline model's encoder were substituted by 1D convolutional layers (kernel size 1 and padding 0), inspired by \cite{Peng2020ECCV}.  Secondly, we replace all custom max-pooling layers (except the last one) with 1D max-pooling layers, and consequently, the DeepSet layers \cite{zaheer2017deep} were adapted to be compatible with 1D max-pooling. However, the last two fully-connected layers of the encoder were preserved, similar to \cite{Occupancy_Networks,Atzmon_2020_CVPR}, for producing both the mean and standard deviation used to sample a latent code that is fed into the decoder. The architecture is shown in Figure~\ref{fig:encoder}.

\begin{figure}[t]
\begin{center}
\centering\includegraphics[width=\columnwidth]{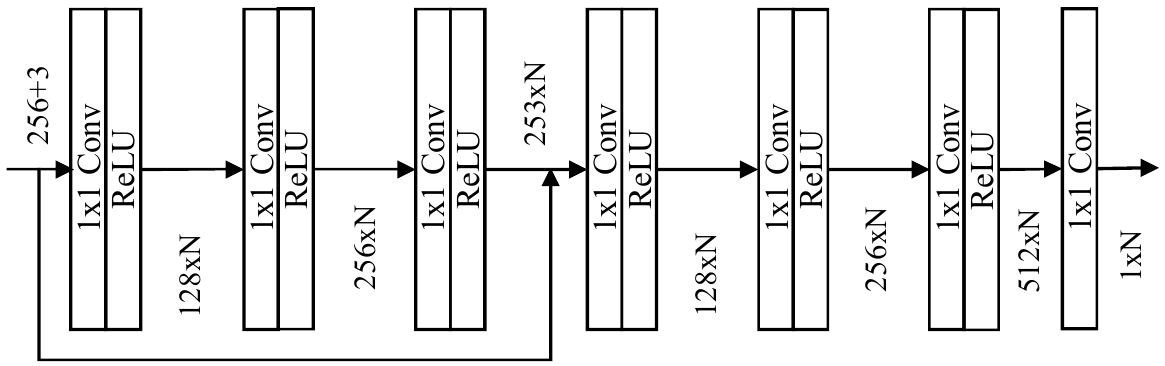}
\caption{The proposed LightSAL decoder network architecture.}
\label{fig:decoder}
\end{center}
\end{figure}

\subsection{Decoder}
\label{Decoder}
The LightSAL decoder consists of six 1D convolutional layers (kernel size 1, padding 0), each convolutional layer followed by a Rectified Linear Unit (ReLU)~\cite{nair2010rectified} activation function. One skip connection has been added between the input and the third layer to propagate the initial raw information to higher layers, for better learning. A similar skip connection is present in the 8-layer DeepSDF architecture \cite{park2019deepsdf} (used by baseline SAL), where the skip connection inserts the latent vector to the network pipeline after the 4th layer. LightSAL does not feature any skip connections between the encoder and the decoder. 

In the DeepSDF decoder, each fully-connected layer has 512 filters, which results in more than 1M trainable parameters. Based on empirical results, we have observed that such a number of parameters does not significantly benefit reconstruction accuracy. Thus, the LightSAL decoder features significantly fewer filters in the pattern $(128\longrightarrow256\longrightarrow512\longrightarrow128\longrightarrow256\longrightarrow512)$. Finally, similar to baseline SAL \cite{Atzmon_2020_CVPR}, the LightSAL decoder does not use an activation layer at the end of the decoder pipeline. This is in contrast to DeepSDF, where a \textit{tanh} layer is used.

\subsection{Training and inference}
\label{ssec:training}
The proposed LightSAL architecture was trained with the Adam optimizer~\cite{kingma2014adam}. The initial learning rate was 0.0005 and a batch size of 16 was used for training each model presented in this paper. The scheduler was set to decrease the learning rate by a factor 0.5 after every 200 epochs. All models were trained with 500 epochs on a single 24GB GeForce RTX 3090 GPU. Each epoch required about 65$\pm$3 seconds on the D-Faust dataset when every 5th training sample\footnote{Because of dense temporal sampling in this dataset and making a fair comparison with baseline.} was used from the full 75\% training dataset of 41k D-Faust samples.

During the inference phase, we used the Marching Cubes algorithm \cite{lorensen1987marching} to mesh the learned implicit representation from the test sample raw scan. For evaluation purposes, 30k points were sampled uniformly from the meshed surface in order to compute the Chamfer distance between the reconstructed and the ground truth shapes.

\section{Experiments}
\label{sec:experiments}
We evaluated the proposed LightSAL architecture on the D-Faust~\cite{dfaust:CVPR:2017} dataset that was also used in \cite{Atzmon_2020_CVPR} for the baseline SAL model. Three different types of training and test procedures were conducted to compare the proposed architecture with the baseline: (a) learning shape space from raw scans, (b) reconstructing \textit{unseen humans}, and (c) reconstructing shapes of \textit{unseen human poses}. This set of experimental procedures (a)-(c) is the same that was used by the baseline SAL \cite{Atzmon_2020_CVPR} work. Besides visual results, the Chamfer distances for all three training procedures are reported.  

\textbf{\underline{Dataset}:}
The D-Faust \cite{dfaust:CVPR:2017} dataset consists of 41k data samples (triangle soup, $X_{i}$) of 10 human subjects performing 129 different actions. The raw D-Faust data contains usual defects such as noise, holes, and occasional artifacts caused by reflections. The same train-test split files as provided with \cite{Atzmon_2020_CVPR} were used to train and test the LightSAL models. The unsigned distance for each sample to the closest triangle was pre-computed using the CGAL library \cite{alliez2010point} for faster training and testing.

\textbf{\underline{Baseline architecture}:}
The SAL neural network architecture presented in \cite{Atzmon_2020_CVPR} was used as the baseline for evaluating LightSAL. Both the baseline architecture, and the proposed architecture are trainable directly from raw scans.  

\subsection{Human shape reconstruction from raw scans}
\label{ssec:firstexperiment}
In human shape reconstruction, the encoder learns the shape representation from the uniformly sampled input point cloud $\textbf{X} \in R^{N\times3}$ (where $N = 128^{2}$) drawn from raw scans $X_{i}$, where the encoder $(\mu,\eta) = g(\textbf{X}; w)$ is represented either by the LightSAL encoder (Section~\ref{encoder}), or the baseline SAL encoder, for comparison purposes. 
Here, $\mu \in R^{256}$  represents the latent vector and $\eta\in R^{256}$ is the diagonal covariance matrix computed by $\Sigma$ = diag exp  $\eta$. Learning from the input point cloud \textcolor{red}{\textbf{X}}, the encoder infers probability values $\mathcal{N}(\mu,\Sigma)$. Consequently, the decoder decodes the implicit representation of the input point cloud with the addition of a latent code. The training was conducted using sign agnostic learning loss with $L^{2}$ distance, as proposed in \cite{Atzmon_2020_CVPR}, combined with variational auto-encoder loss \cite{kingma2013auto}.

In the inference phase, the reconstructed surfaces were generated by a forward pass of a point cloud sampled uniformly at random from raw unseen test scans. The implicit representation, yielding zero level-sets of an implicit function of the input point cloud, was meshed and Chamfer distance was computed to estimate the quantitative performance of the proposed approach compared to the baseline.      

We did not re-train the baseline model for this experiment\footnote{Considerable effort was invested in trying to train the baseline SAL model with batch size 16, but the resulting model never yielded high-quality results.}, instead we relied on the 2000-epoch pretrained model provided\footnote {https://github.com/matanatz/SAL} by the baseline work authors for reconstructing the shapes and to estimate the Chamfer distances.

\begin{figure*}
\begin{center}
\begin{adjustbox}{max size={\textwidth}{\textheight}}
\begin{tabular}{c c c c c c c c c c c c c c}
 \includegraphics[ height= 0.7 in]{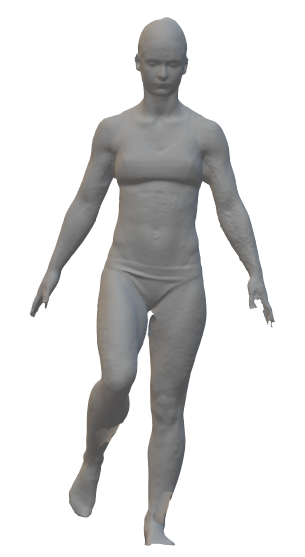}&\includegraphics[ height= 0.7 in]{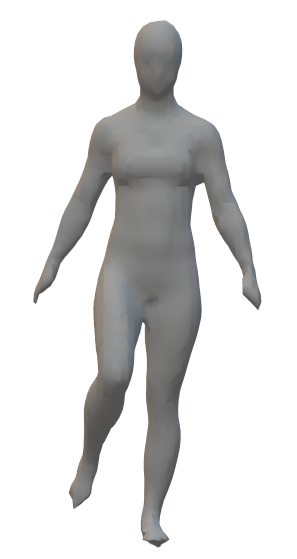}&\includegraphics[height= 0.7 in]{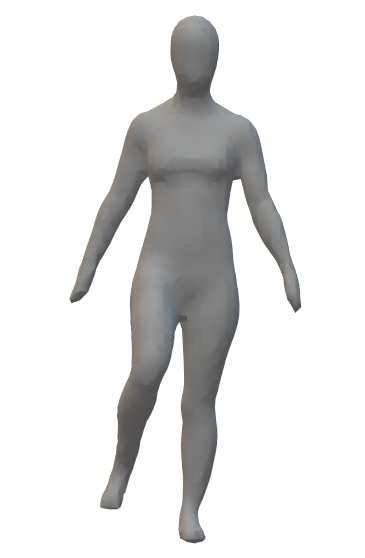}&\includegraphics[height= 0.7 in]{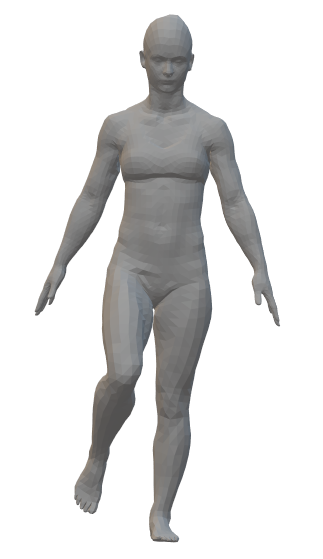}&&\includegraphics[height= 0.7 in]{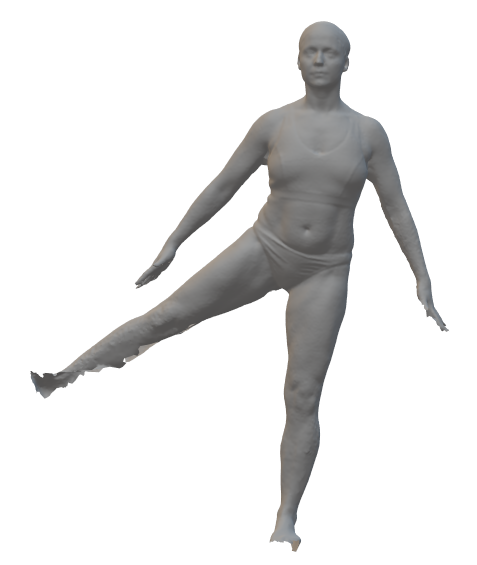}&\includegraphics[height= 0.7 in]{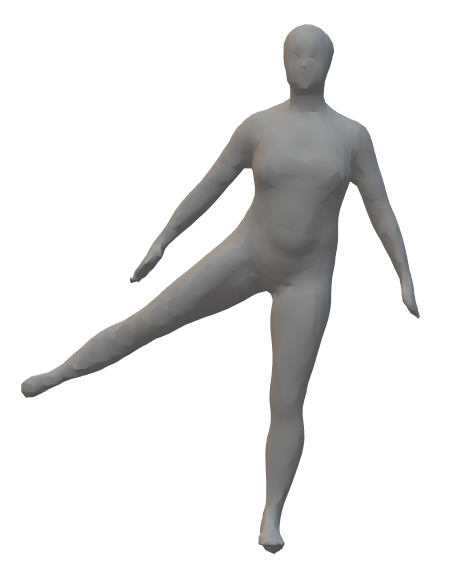}&\includegraphics[height= 0.7 in]{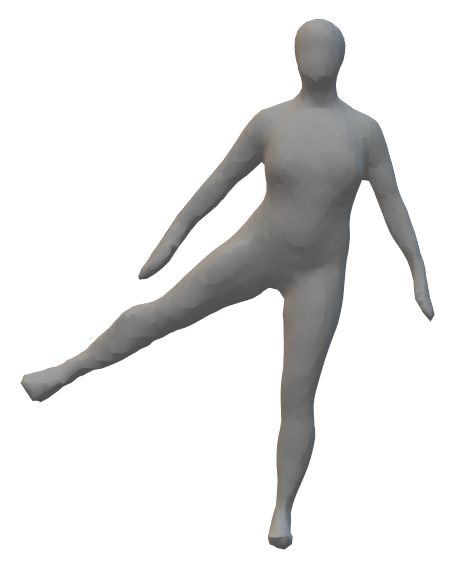}&\includegraphics[height= 0.7 in]{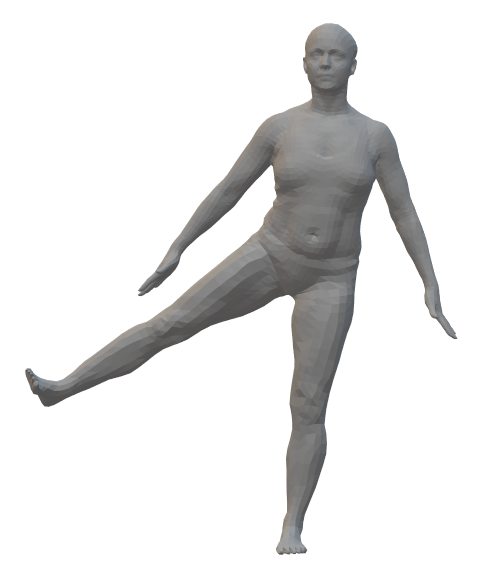}&&\includegraphics[height= 0.7 in]{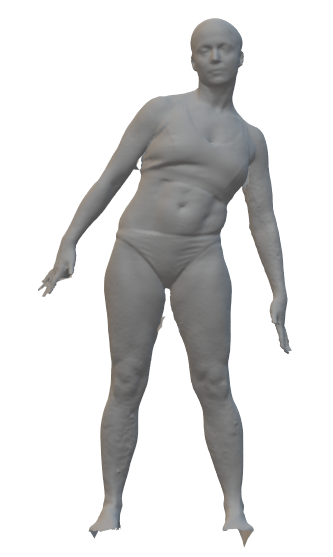}&\includegraphics[height= 0.7 in]{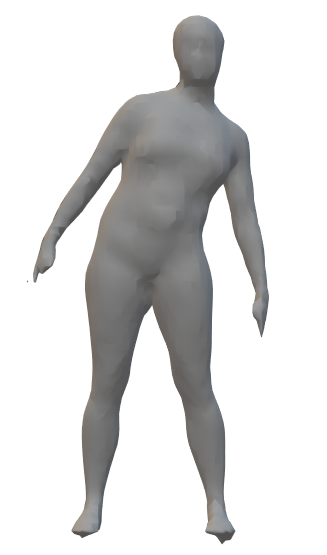}&\includegraphics[height= 0.7 in]{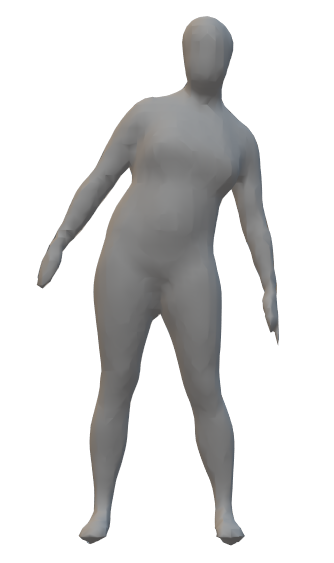}&\includegraphics[height= 0.7 in]{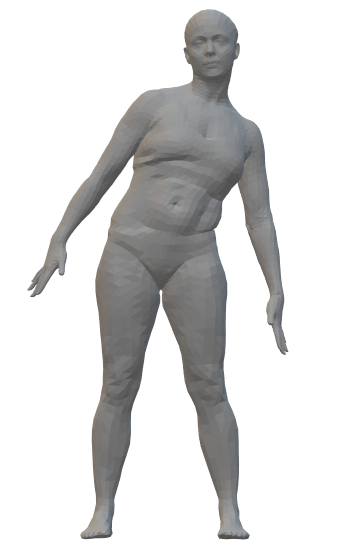}\\
\includegraphics[height= 0.7 in]{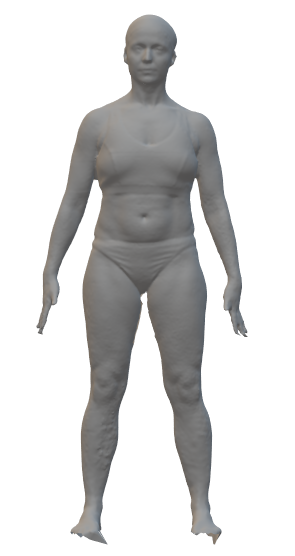}&\includegraphics[ height= 0.7 in]{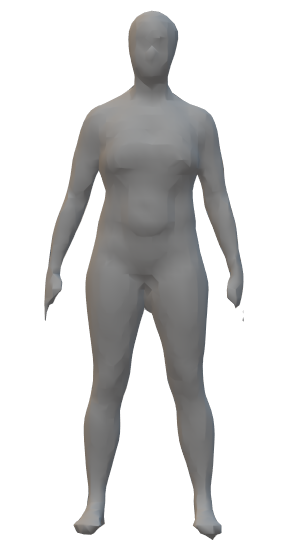}&\includegraphics[height= 0.7 in]{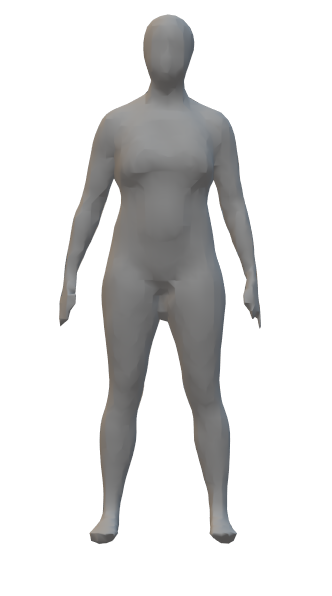}&\includegraphics[height= 0.7 in]{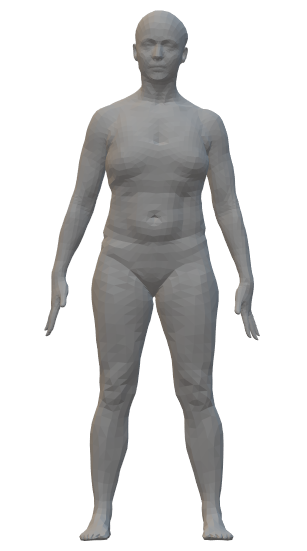}&&\includegraphics[height= 0.7 in]{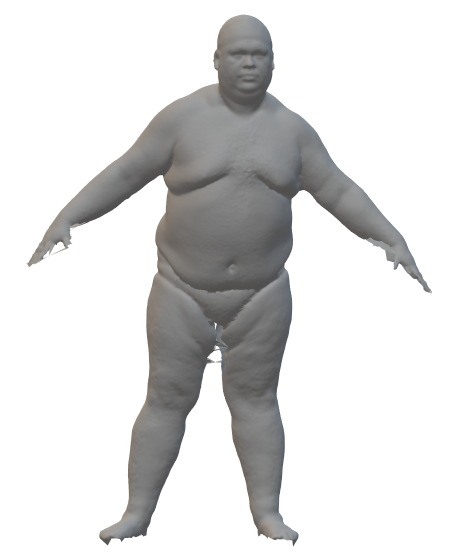}&\includegraphics[height= 0.7 in]{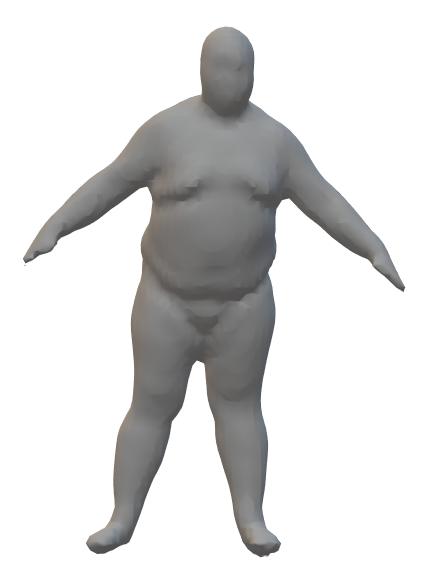}&\includegraphics[height= 0.7 in]{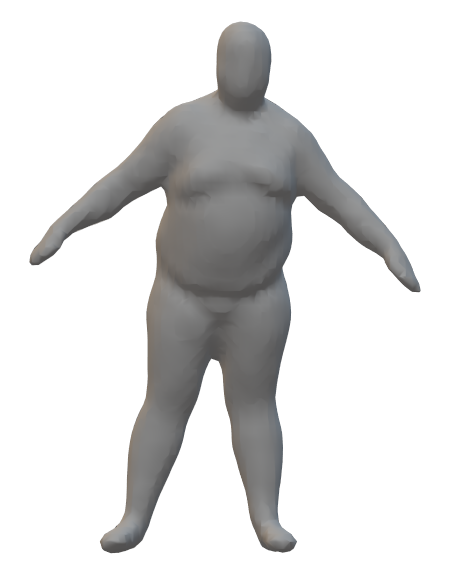}&\includegraphics[height= 0.7 in]{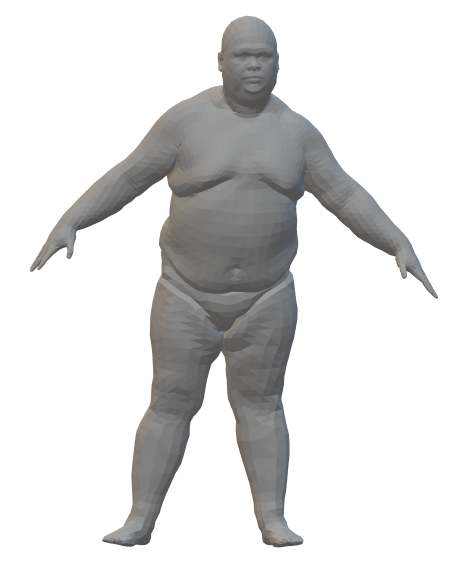}&&\includegraphics[height= 0.7 in]{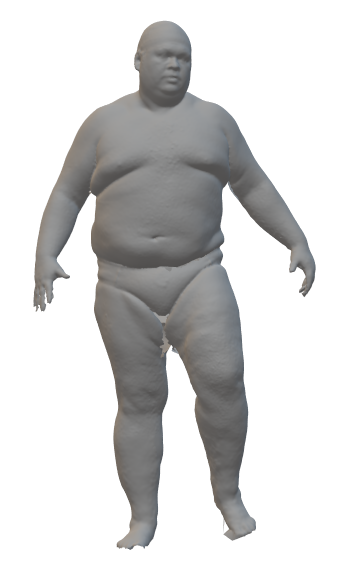}&\includegraphics[ height= 0.7 in]{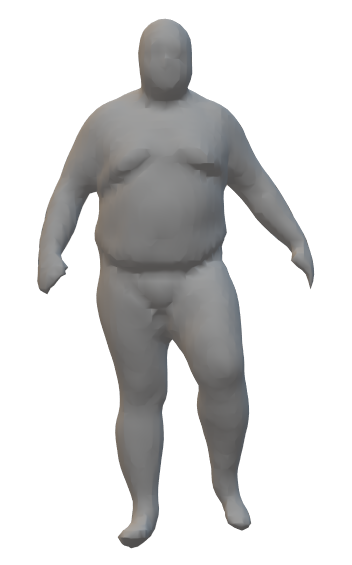}&\includegraphics[ height= 0.7 in]{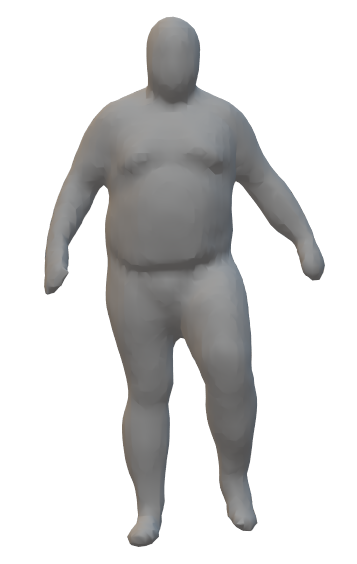}&\includegraphics[ height= 0.7 in]{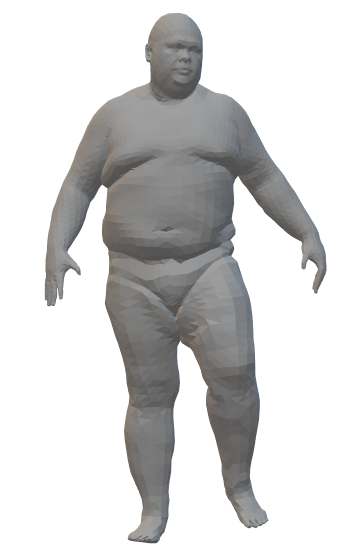}\\
\includegraphics[height= 0.7 in]{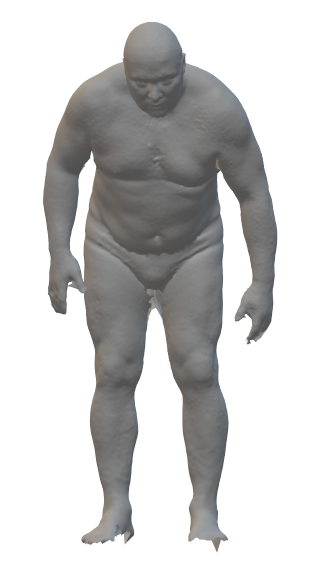}&\includegraphics[height= 0.7 in]{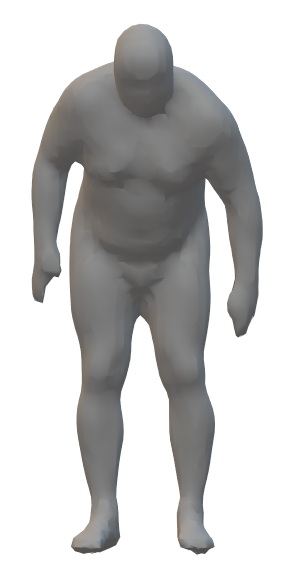}&\includegraphics[ height= 0.7 in]{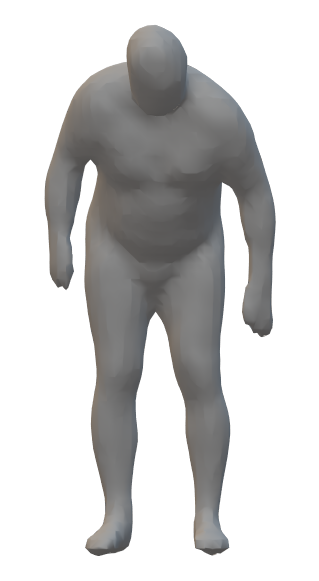}&\includegraphics[ height= 0.7 in]{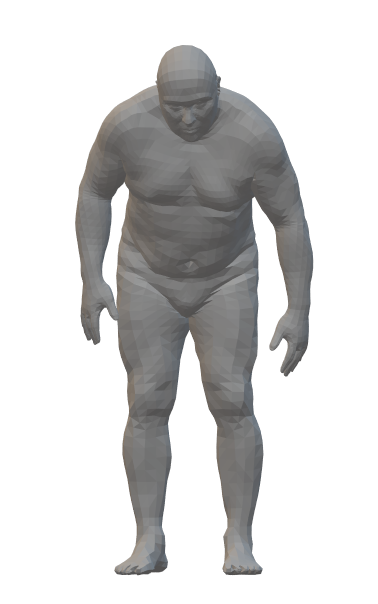}&&\includegraphics[ height= 0.7 in]{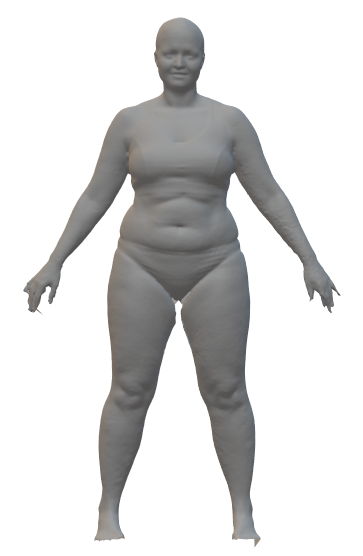}&\includegraphics[ height= 0.7 in]{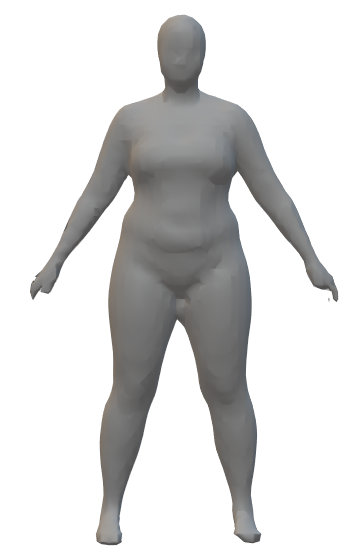}&\includegraphics[ height= 0.7 in]{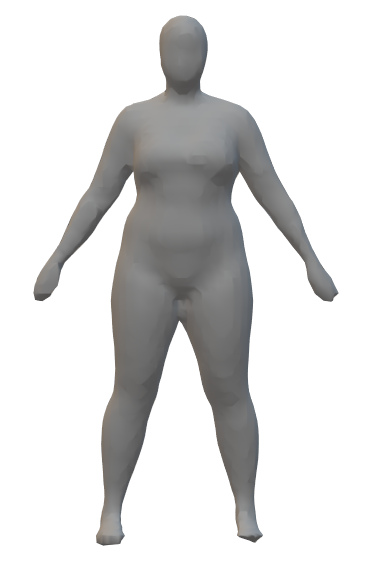}&\includegraphics[height= 0.7 in]{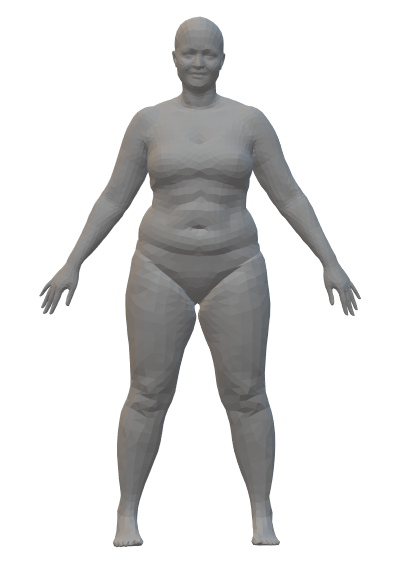}&&\includegraphics[height= 0.7 in]{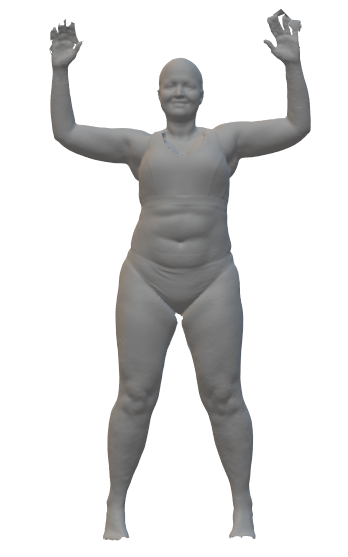}&\includegraphics[height= 0.7 in]{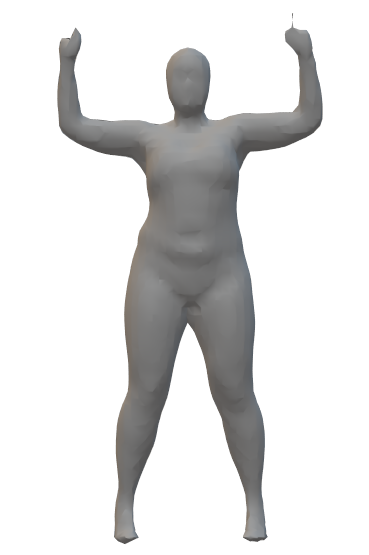}&\includegraphics[height= 0.7 in]{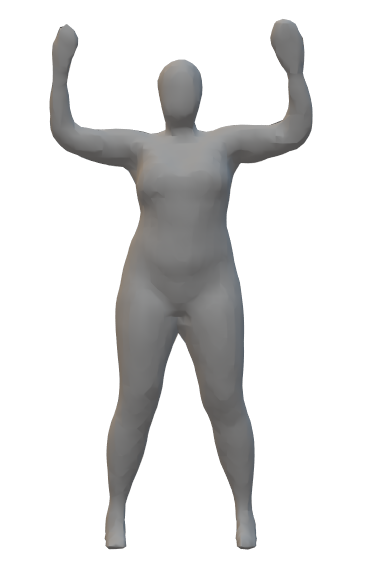}&\includegraphics[height= 0.7 in]{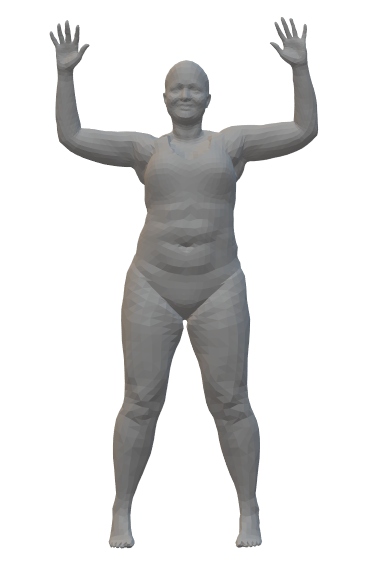}\\
\includegraphics[height= 0.7 in]{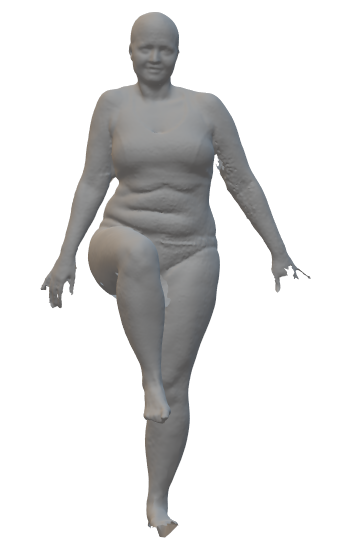}&\includegraphics[height= 0.7 in]{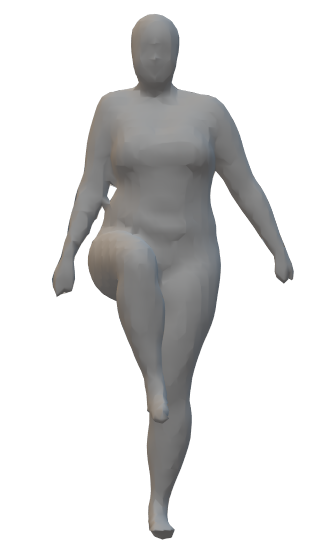}&\includegraphics[height= 0.7 in]{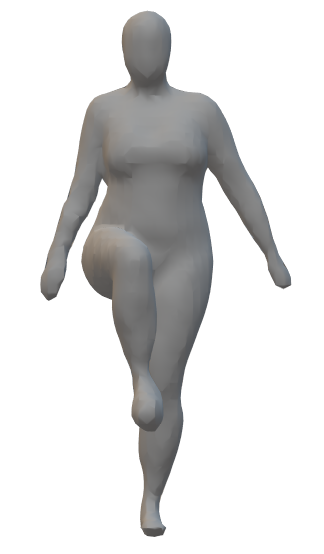}&\includegraphics[height= 0.7 in]{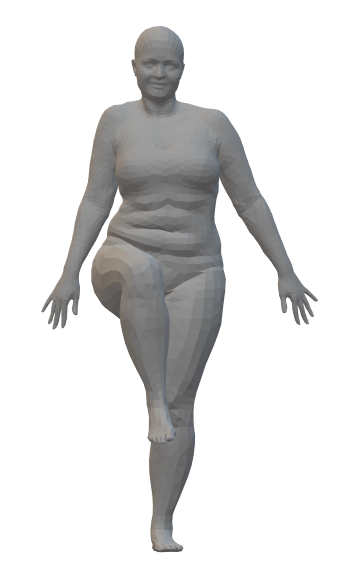}&&\includegraphics[height= 0.7 in]{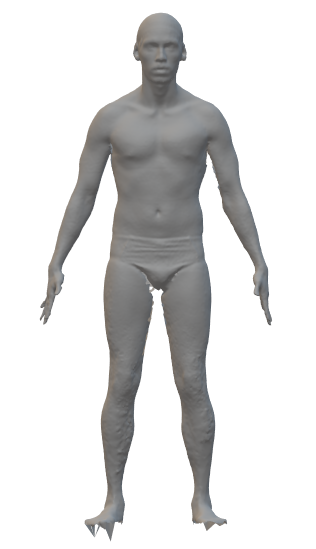}&\includegraphics[height= 0.7 in]{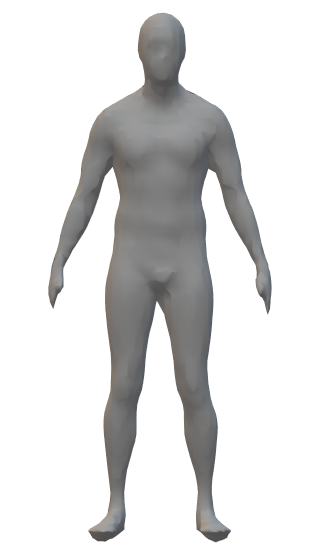}&\includegraphics[height= 0.7 in]{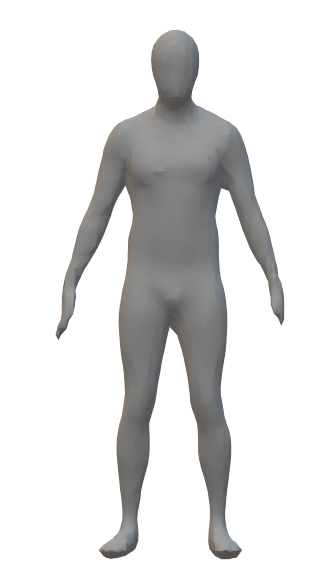}&\includegraphics[height= 0.7 in]{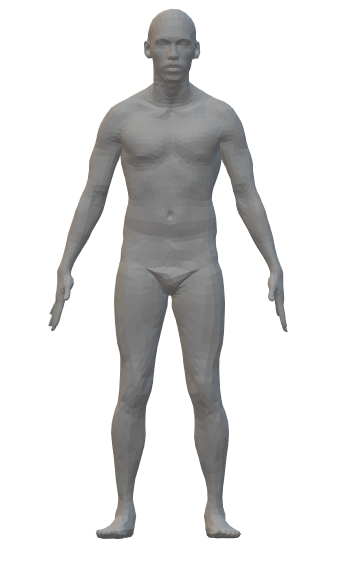}&&\includegraphics[height= 0.7 in]{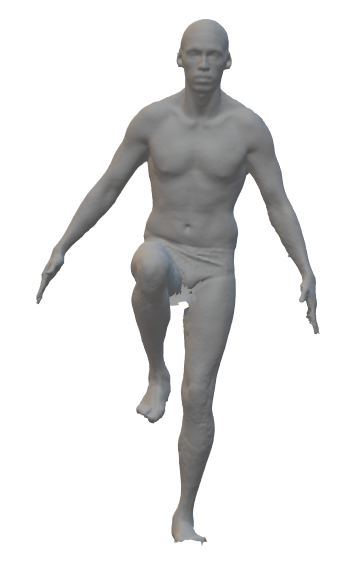}&\includegraphics[height= 0.7 in]{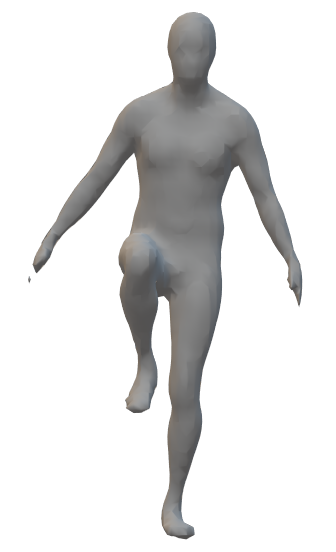}&\includegraphics[height= 0.7 in]{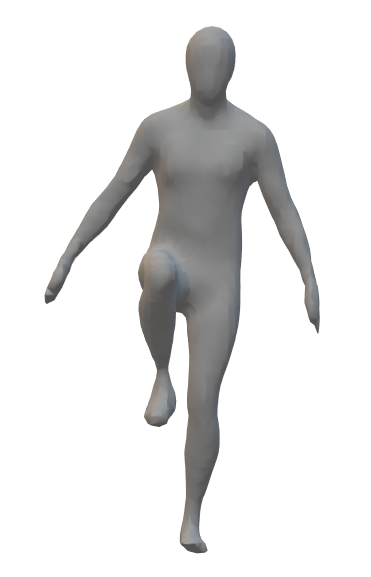}&\includegraphics[height= 0.7 in]{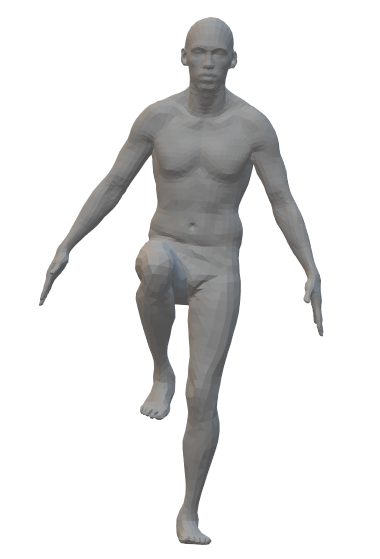}\\
\end{tabular}
\end{adjustbox}
\end{center}
   \caption{Reconstructed human shapes from the D-Faust dataset. Each row shows three different cases, and each case consists of four subfigures: (from left to right) input test scan, baseline SAL, LightSAL, ground truth. Here, baseline SAL has been trained with 2000 epochs, whereas LightSAL has been trained with 500 epochs.}
\label{Fig.1:reconstructed_human_shape:Dfaust_dataset_primary}
\end{figure*}

The quantitative and qualitative results of LightSAL and the baseline for this experiment are shown in Table~\ref{chamfer_distance:Dfaust_Dataset_primary} and Figure~\ref{Fig.1:reconstructed_human_shape:Dfaust_dataset_primary}, respectively. Similar to \cite{Atzmon_2020_CVPR}, we report both train and test time performance. In contrast to unseen human and unseen pose experiments (Section~\ref{ssec:unseen}), the LightSAL architecture outperformed the baseline architecture in terms of Chamfer distance (Table~\ref{chamfer_distance:Dfaust_Dataset_primary}) only in one case, otherwise coming close to the quality of SAL. Visually, the LightSAL and SAL results are close to each other (Figure~\ref{Fig.1:reconstructed_human_shape:Dfaust_dataset_primary}).

\begin{table}[t]
\begin{center}
\begin{adjustbox}{width=\columnwidth}
\begin{tabular}{|c|c|c|c|c|c|c|c|} 
\multicolumn{8}{c}{} \\ \cline{1-8}
\multirow{2}{*}{Type}&\multirow{2}{*}{Method}&\multicolumn{3}{c}{Registrations}&\multicolumn{3}{|c|}{Scans} \\ \cline{3-8}
&&5\%&50\%&95\%&5\%&50\%&95\% \\ \cline{1-8}
\multicolumn{1}{|c|}{Train}&SAL \cite{Atzmon_2020_CVPR}&0.07&0.10&0.20&0.05&0.07&0.09 \\ \cline{2-8}
\multicolumn{1}{|c}{}&\multicolumn{1}{|c|}{Ours}&0.08&0.14&0.31&0.06&0.09&0.14\\ \cline{1-8}
\multicolumn{1}{|c|}{Test}&SAL \cite{Atzmon_2020_CVPR}&0.07&0.12&0.44&0.05&0.07&0.14 \\ \cline{2-8}
\multicolumn{1}{|c}{}&\multicolumn{1}{|c|}{Ours}&0.09&0.15&\textbf{0.42}&0.06&0.09&0.16\\ \cline{1-8}
\multicolumn{8}{c}{}
\end{tabular}
\end{adjustbox}
\caption{Chamfer distances of the reconstructed human shapes against ground truth registrations, and raw scans. The data are presented in percentiles ($5{^{th}}$, $50{^{th}}$, and  $95{^{th}}$), values being multiplied by $10{^3}$. Here, our LightSAL model has been trained with 500 epochs, whereas baseline SAL has been trained by 2000 epochs.}
\label{chamfer_distance:Dfaust_Dataset_primary}
\end{center}
\end{table}

\begin{table}
\begin{center}
\begin{adjustbox}{width=\columnwidth}
\begin{tabular}{|c|c|c|c|c|c|c|c|} 
\multicolumn{8}{c}{} \\ \cline{1-8}
\multirow{2}{*}{Type}&\multirow{2}{*}{Method}&\multicolumn{3}{c}{Registrations}&\multicolumn{3}{|c|}{Scans} \\ \cline{3-8}
&&5\%&50\%&95\%&5\%&50\%&95\% \\ \cline{1-8}
\multicolumn{1}{|c|}{Train}&SAL \cite{Atzmon_2020_CVPR}&0.06&0.09&0.18&0.04&0.06&0.09  \\ \cline{2-8}
\multicolumn{1}{|c}{}&\multicolumn{1}{|c|}{Ours}&0.09&0.14&0.29&0.06&0.08&0.13\\ \cline{1-8}
\multicolumn{1}{|c|}{Test}&SAL \cite{Atzmon_2020_CVPR}&0.26&0.75&4.99&0.14&0.34&1.53 \\ \cline{2-8}
\multicolumn{1}{|c}{}&\multicolumn{1}{|c|}{Ours}&\textbf{0.16}&\textbf{0.34}&\textbf{3.15}&\textbf{0.09}&\textbf{0.17}&\textbf{0.71}\\ \cline{1-8}
\multicolumn{8}{c}{}
\end{tabular}
\end{adjustbox}
\caption{Chamfer distances of reconstructed \textit{unseen humans} against ground truth registrations and raw scans. The data are presented in percentiles ($5{^{th}}$, $50{^{th}}$, and  $95{^{th}}$), values being multiplied by $10{^3}$. Here, our LightSAL model has been trained with 500 epochs, whereas baseline SAL has been trained by 2000 epochs.}
\label{chamfer_distance:Dfaust_Dataset_unseenHuman}
\end{center}
\end{table}
\begin{table}
\begin{center}
\begin{adjustbox}{width=\columnwidth}
\begin{tabular}{|c|c|c|c|c|c|c|c|} 
\multicolumn{8}{c}{} \\ \cline{1-8}
\multirow{2}{*}{Type}&\multirow{2}{*}{Method}&\multicolumn{3}{c}{Registrations}&\multicolumn{3}{|c|}{Scans} \\ \cline{3-8}
&&5\%&50\%&95\%&5\%&50\%&95\% \\ \cline{1-8}
\multicolumn{1}{|c|}{Train}&SAL \cite{Atzmon_2020_CVPR}&0.08&0.12&0.25&0.05&0.07&0.1 \\ \cline{2-8}
\multicolumn{1}{|c}{}&\multicolumn{1}{|c|}{Ours}&0.09&0.14&0.29&0.06&0.08&0.13\\ \cline{1-8}
\multicolumn{1}{|c|}{Test}&SAL \cite{Atzmon_2020_CVPR}&0.11&0.37&2.26&0.07&0.18&0.93 \\ \cline{2-8}
\multicolumn{1}{|c}{}&\multicolumn{1}{|c|}{Ours}&\textbf{0.09}&\textbf{0.19}&\textbf{1.06}&\textbf{0.06}&\textbf{0.11}&\textbf{0.31}\\ \cline{1-8}
\multicolumn{8}{c}{}
\end{tabular}
\end{adjustbox}
\caption{Chamfer distances of reconstructed \textit{unseen human poses} against ground truth registrations and raw scans. Results presented in percentiles ($5{^{th}}$, $50{^{th}}$, and  $95{^{th}}$), values being multiplied by $10{^3}$. Here, our LightSAL model has been trained with 500 epochs, whereas baseline SAL has been trained by 2000 epochs.}
\label{chamfer_distance:Dfaust_Dataset_unseenPose}
\end{center}
\end{table}

\subsection{Generalization to unseen data}
\label{ssec:unseen}

In this experiment, two different cases are covered: (a) reconstructing unseen humans, and (b) reconstructing unseen human poses. For (a), out of the 10 human shape (5 male and 5 female) classes of the D-Faust dataset, 8 classes (4 male and 4 female) were used for training, and 2 classes (1 male and 1 female) were left for testing. On the other hand, for (b), randomly selected two human poses from each human class were left out for testing and the rest of the data were used to train the proposed neural network. The models trained for unseen human shapes and unseen human poses were not provided by the authors of the baseline SAL work, for which reason the numbers appearing in Table~\ref{chamfer_distance:Dfaust_Dataset_unseenHuman} and Table~\ref{chamfer_distance:Dfaust_Dataset_unseenPose} are adopted directly from the SAL publication \cite{Atzmon_2020_CVPR}.
\textit{In unseen human shape generation, LightSAL outperformed the baseline in test-time reconstruction in all cases} (Table~\ref{chamfer_distance:Dfaust_Dataset_unseenHuman} and Table~\ref{chamfer_distance:Dfaust_Dataset_unseenPose}). 

\begin{figure*}
\begin{center}
\begin{adjustbox}{max size={\textwidth}{\textheight}}
\begin{tabular}{c c c c c c c c c c c c c c}
\includegraphics[ height= 0.7 in]{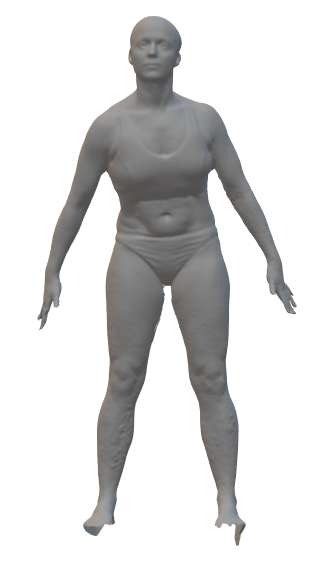}&\includegraphics[height= 0.7 in]{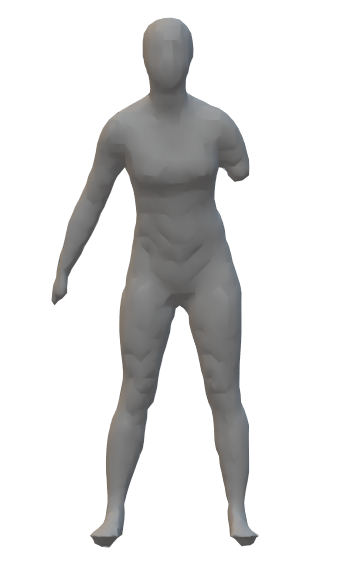}&\includegraphics[height= 0.7 in]{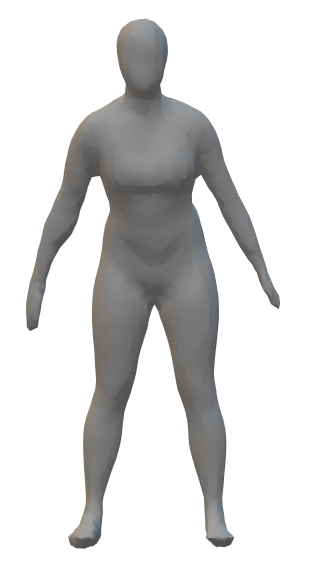}&\includegraphics[height= 0.7 in]{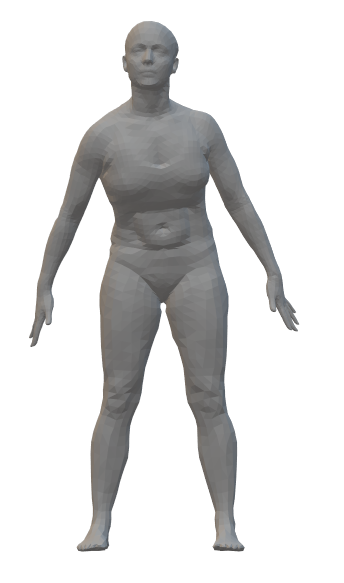}&&\includegraphics[height= 0.7 in]{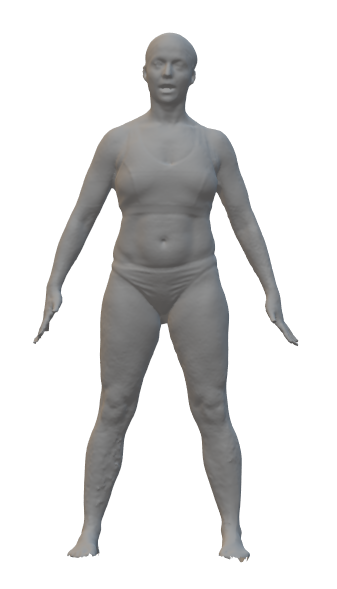}&\includegraphics[height= 0.7 in]{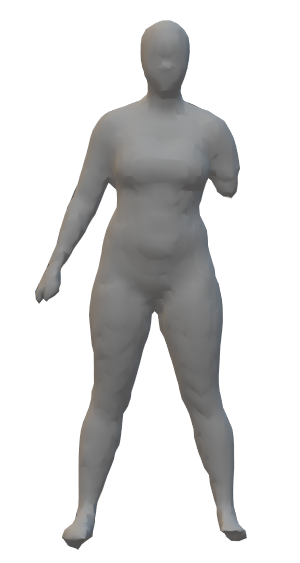}&\includegraphics[height= 0.7 in]{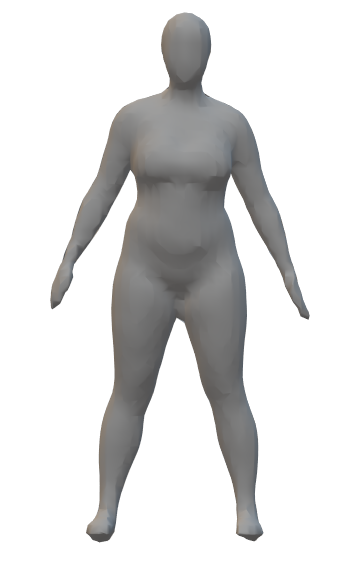}&\includegraphics[height= 0.7 in]{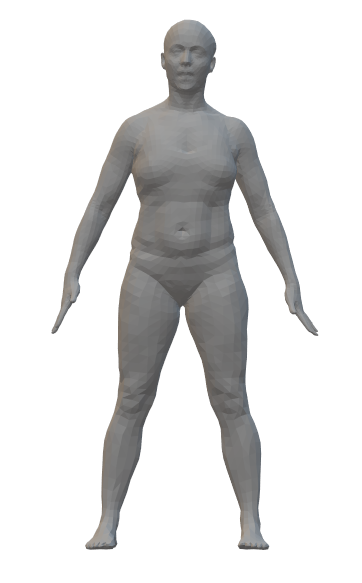}&&\includegraphics[height= 0.7 in]{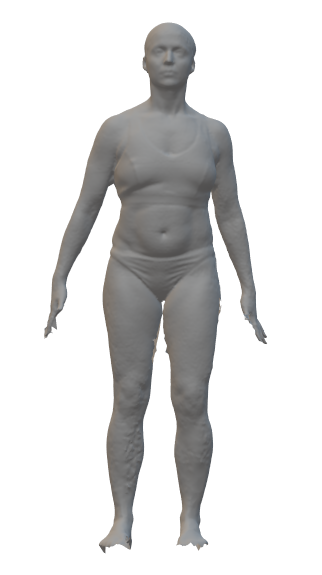}&\includegraphics[height= 0.7 in]{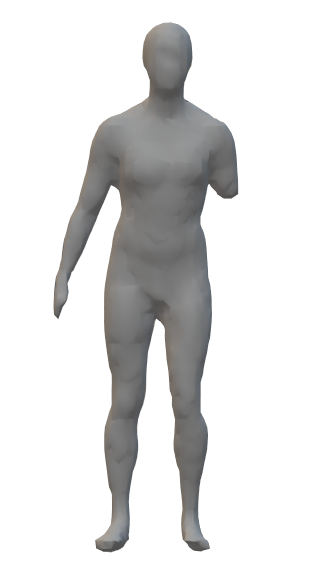}&\includegraphics[height= 0.7 in]{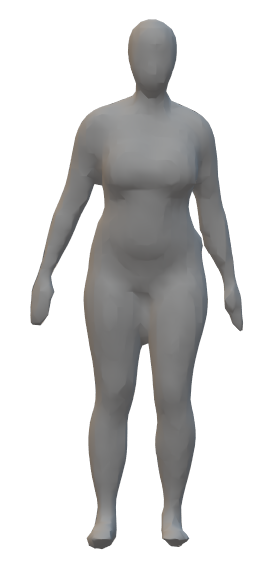}&\includegraphics[height= 0.7 in]{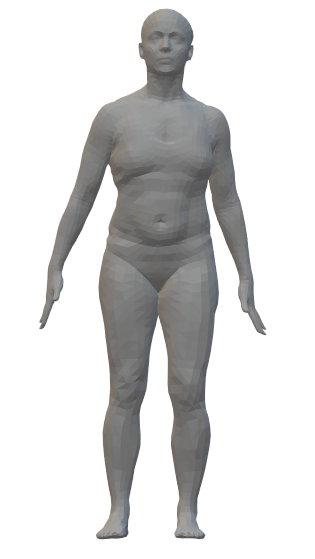}\\
\includegraphics[height= 0.7 in]{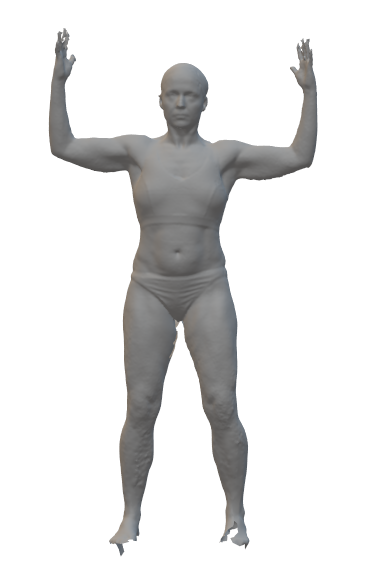}&\includegraphics[height= 0.7 in]{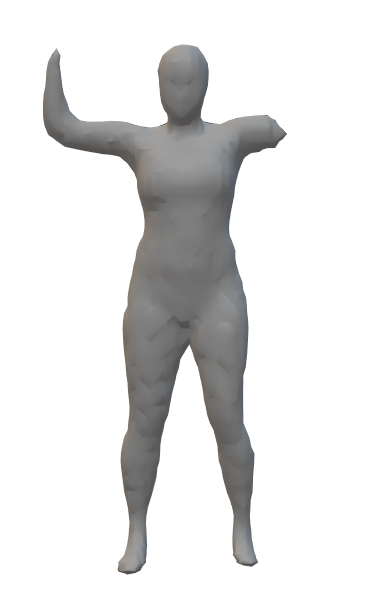}&\includegraphics[height= 0.7 in]{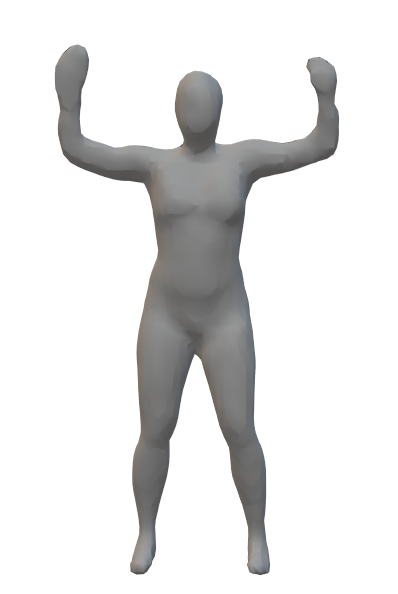}&\includegraphics[height= 0.7 in]{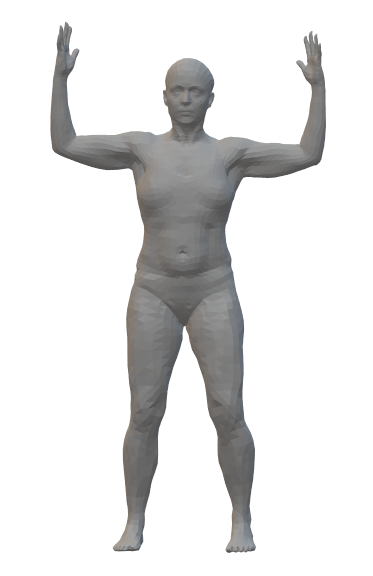}&&\includegraphics[height= 0.7 in]{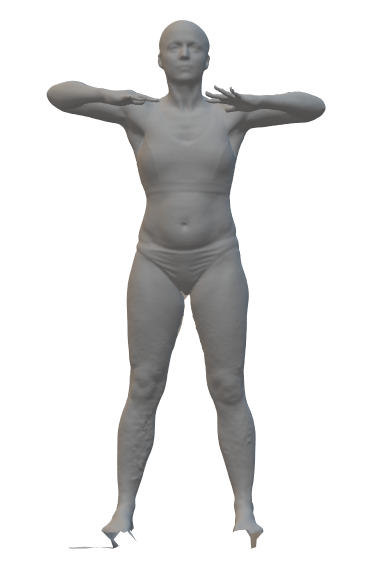}&\includegraphics[height= 0.7 in]{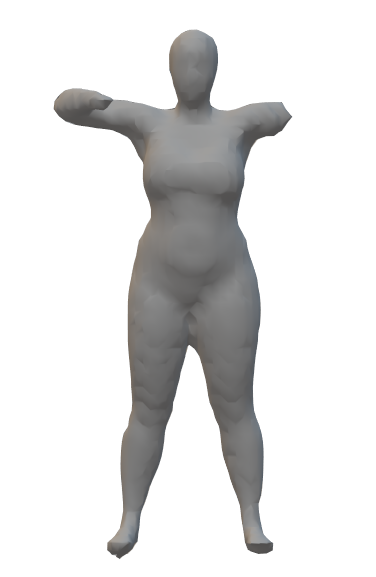}&\includegraphics[height= 0.7 in]{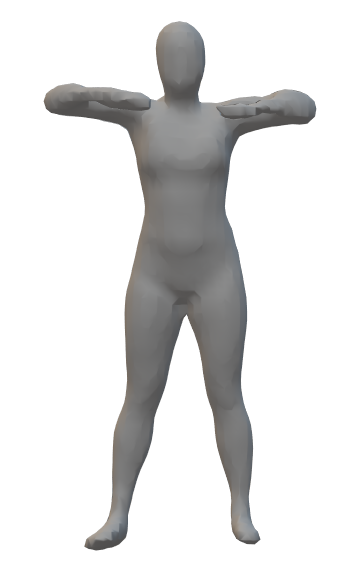}&\includegraphics[height= 0.7 in]{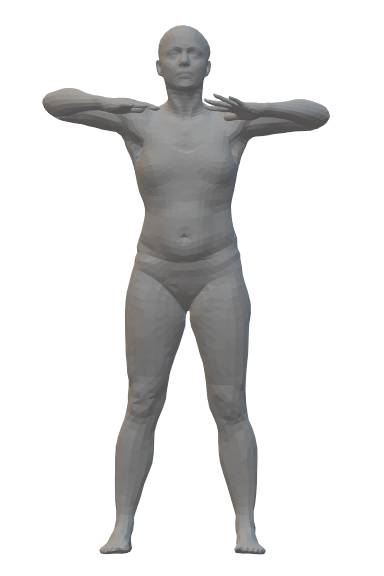}&&\includegraphics[height= 0.7 in]{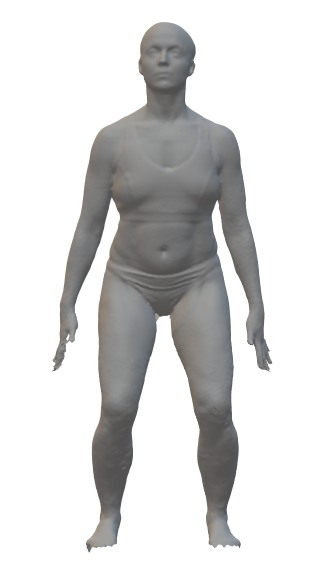}&\includegraphics[height= 0.7 in]{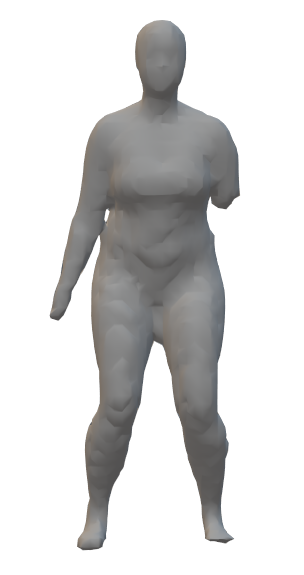}&\includegraphics[height= 0.7 in]{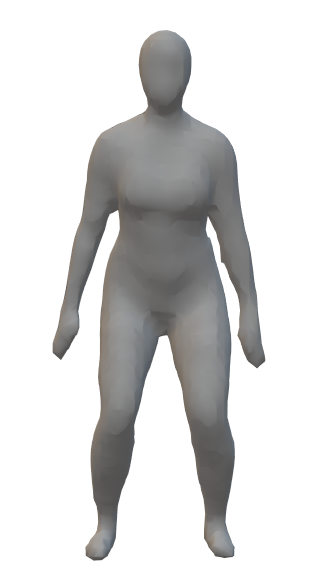}&\includegraphics[height= 0.7 in]{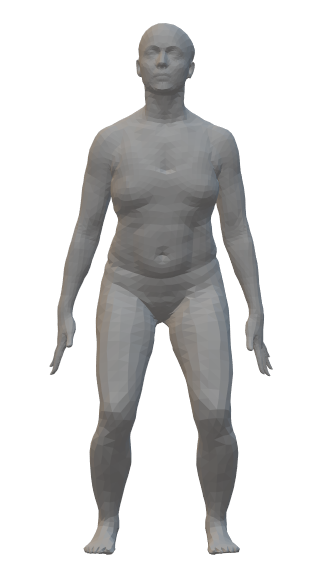}\\
\includegraphics[height= 0.7 in]{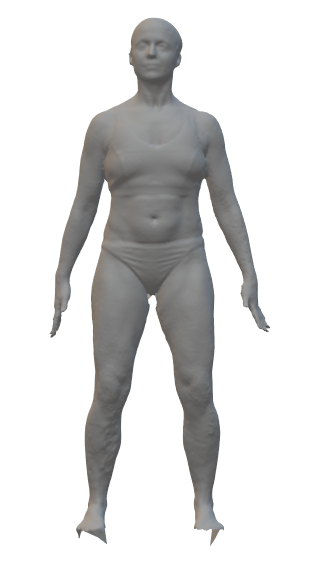}&\includegraphics[height= 0.7 in]{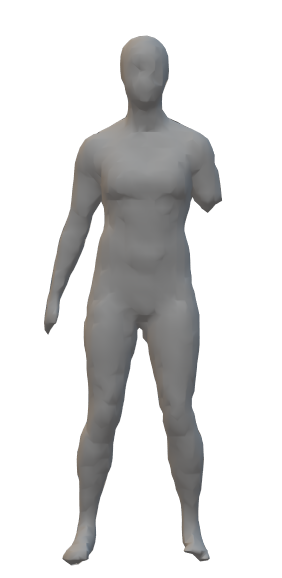}&\includegraphics[height= 0.7 in]{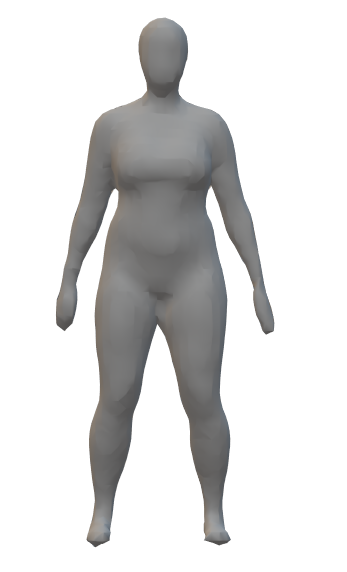}&\includegraphics[height= 0.7 in]{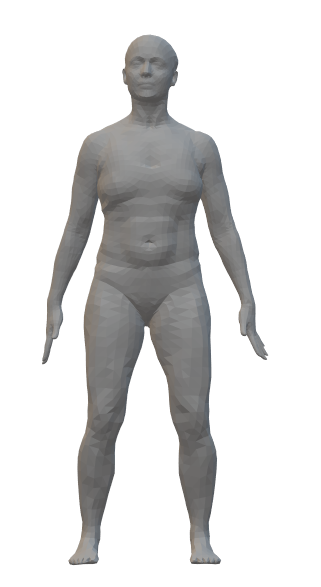}&&\includegraphics[height= 0.7 in]{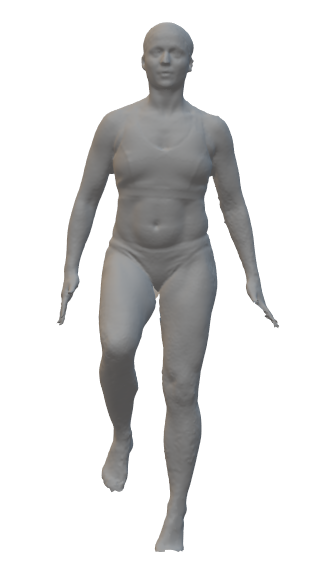}&\includegraphics[height= 0.7 in]{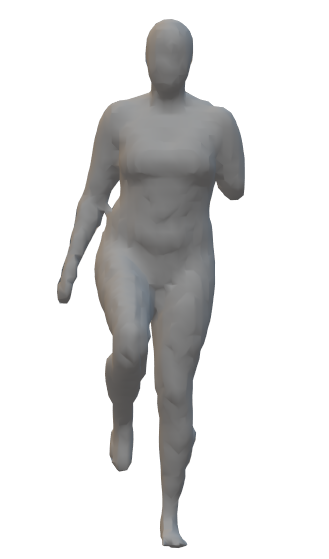}&\includegraphics[height= 0.7 in]{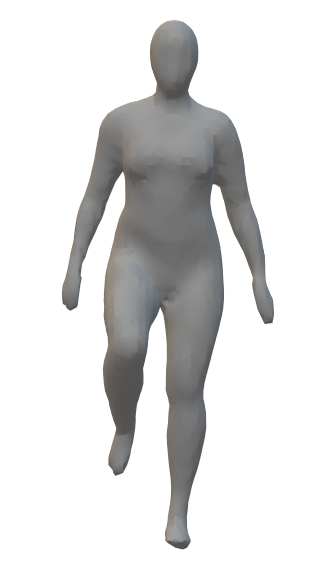}&\includegraphics[height= 0.7 in]{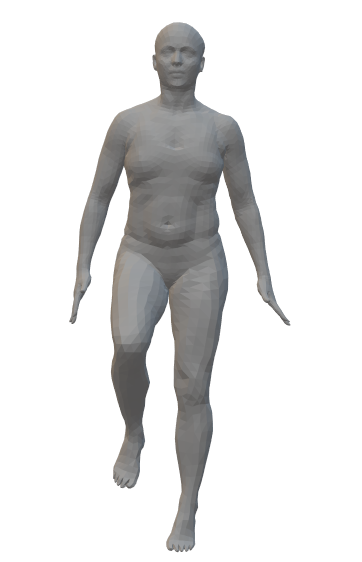}&&\includegraphics[height= 0.7 in]{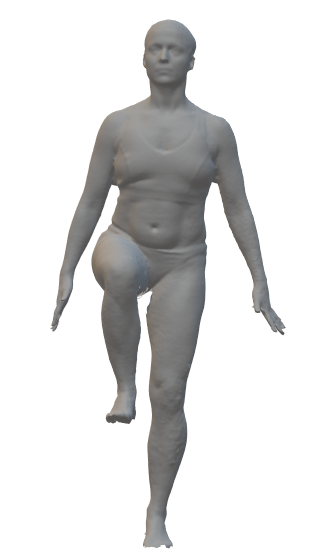}&\includegraphics[height= 0.7 in]{f2f_sl_50025_one_leg_loose.000892.png}&\includegraphics[height= 0.7 in]{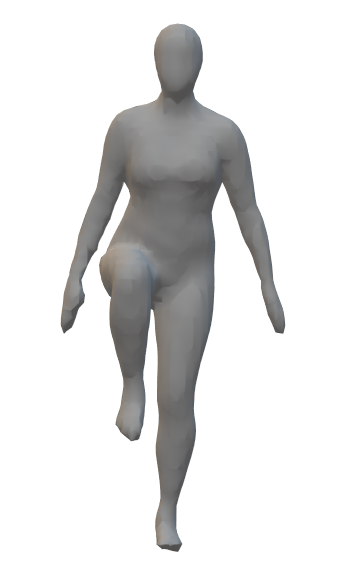}&\includegraphics[height= 0.7 in]{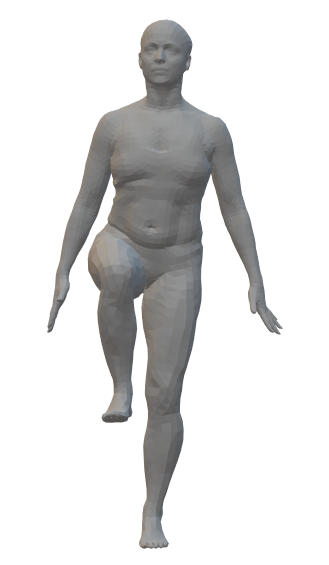}\\
\includegraphics[height= 0.7 in]{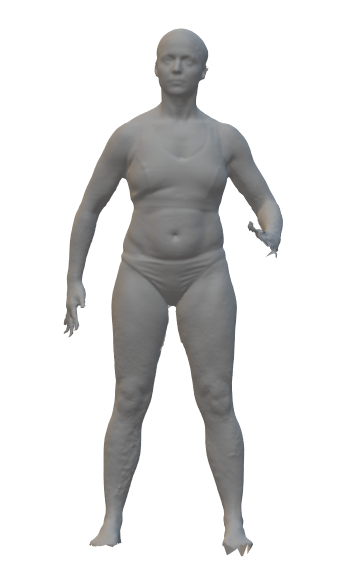}&\includegraphics[height= 0.7 in]{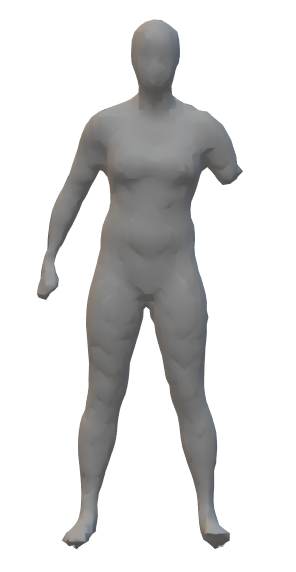}&\includegraphics[height= 0.7 in]{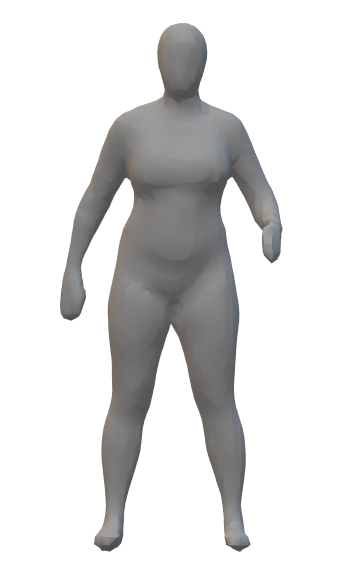}&\includegraphics[height= 0.7 in]{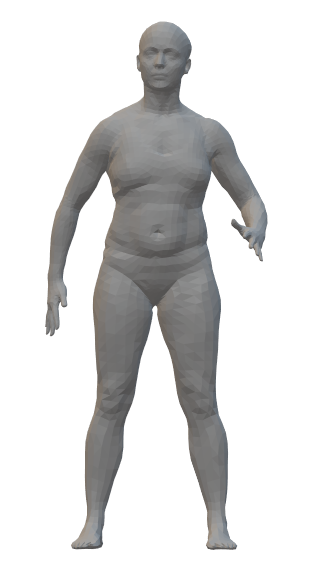}&&\includegraphics[height= 0.7 in]{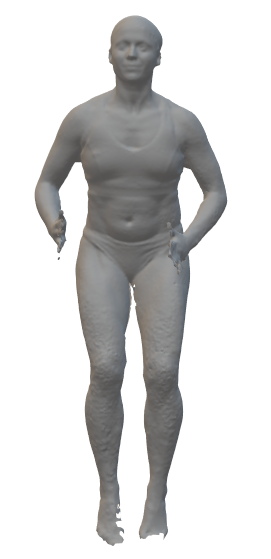}&\includegraphics[height= 0.7 in]{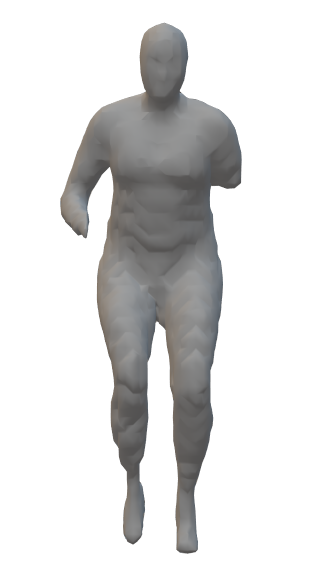}&\includegraphics[height= 0.7 in]{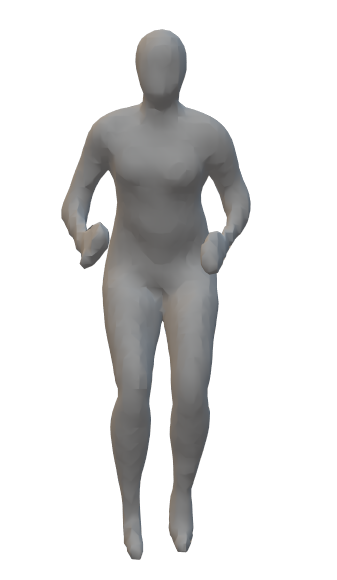}&\includegraphics[height= 0.7 in]{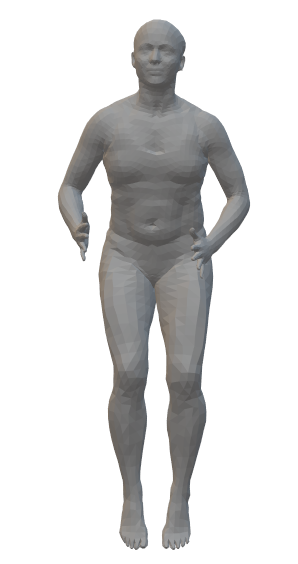}&&\includegraphics[height= 0.7 in]{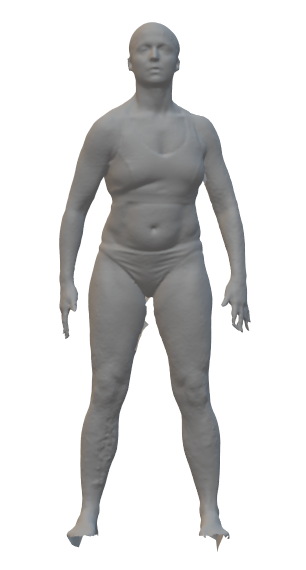}&\includegraphics[height= 0.7 in]{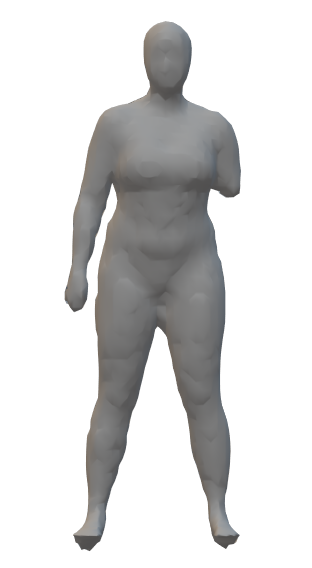}&\includegraphics[height= 0.7 in]{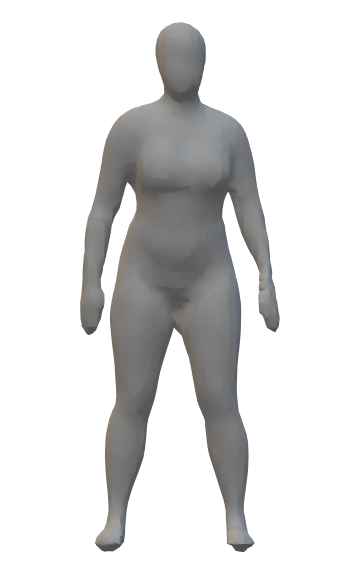}&\includegraphics[height= 0.7 in]{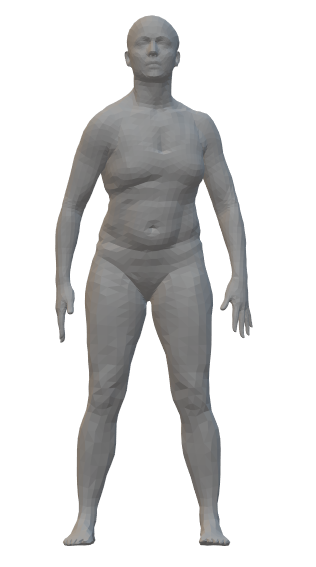}\\
\end{tabular}
\end{adjustbox}
\end{center}
   \caption{The reconstructed unseen human (female) shapes from the D-Faust dataset. Each row consists of three cases with four subfigures: (from left to right) input test scan, reconstruction with 500 epochs by baseline architecture, reconstruction with 500 epochs by proposed architecture, ground truth.}
\label{reconstructed_human_shape2:Dfaust_dataset_unseenHuman_female}
\end{figure*}

\begin{figure*}
\begin{center}
\begin{adjustbox}{max size={\textwidth}{\textheight}}
\begin{tabular}{c c c c c c c c c c c c c c}
\includegraphics[height= 0.7 in]{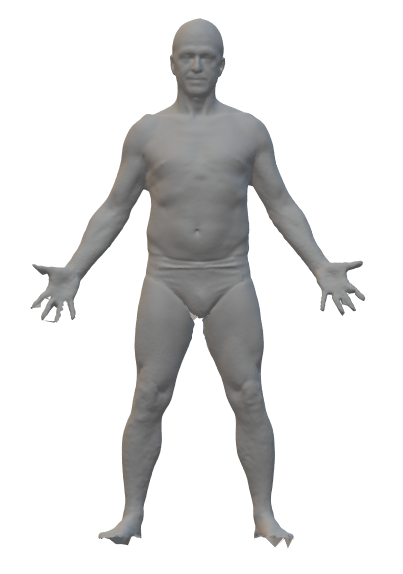}&\includegraphics[height= 0.7 in]{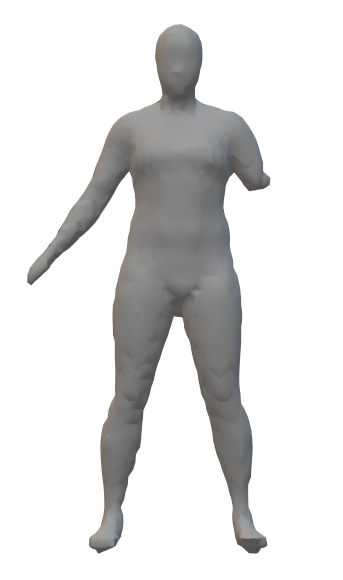}&\includegraphics[height= 0.7 in]{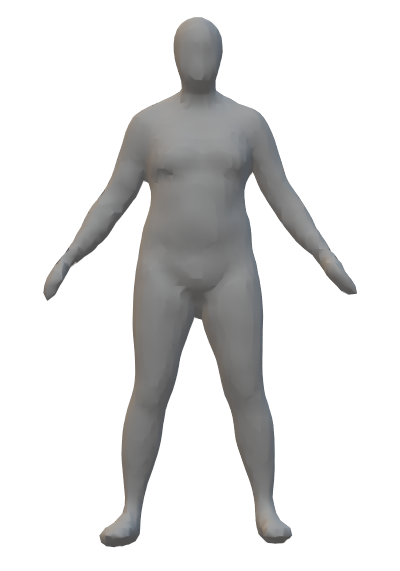}&\includegraphics[height= 0.7 in]{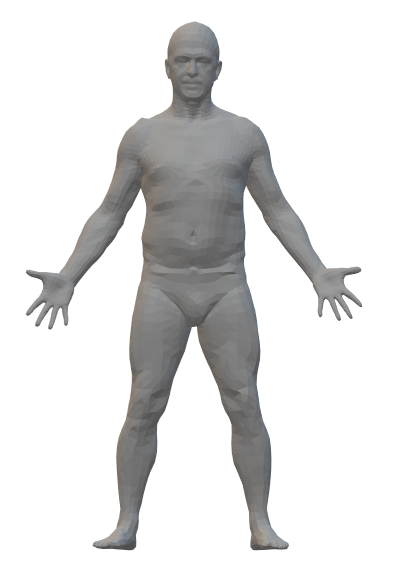}&&\includegraphics[height= 0.7 in]{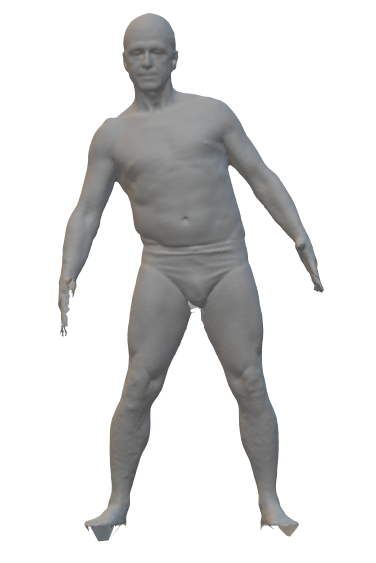}&\includegraphics[height= 0.7 in]{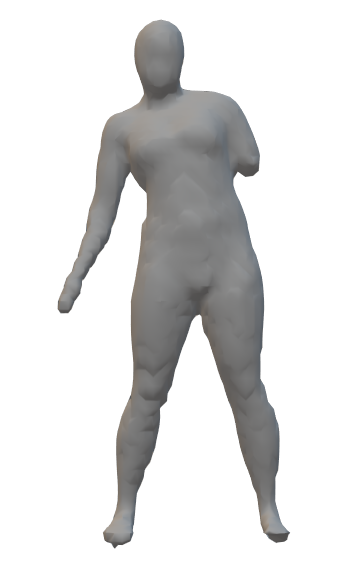}&\includegraphics[height= 0.7 in]{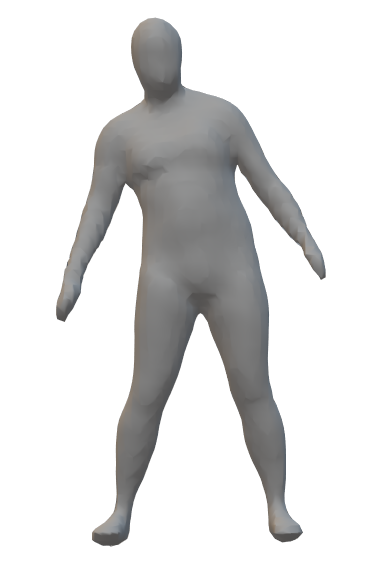}&\includegraphics[height= 0.7 in]{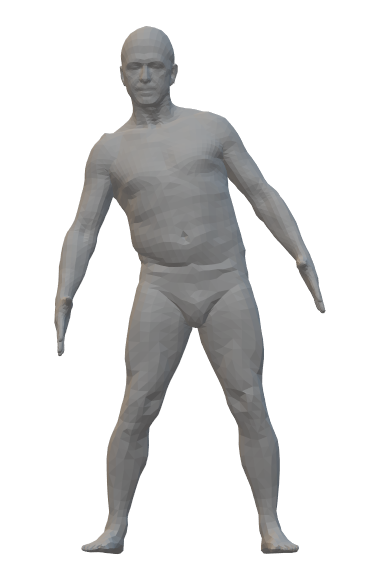}&&\includegraphics[height= 0.7 in]{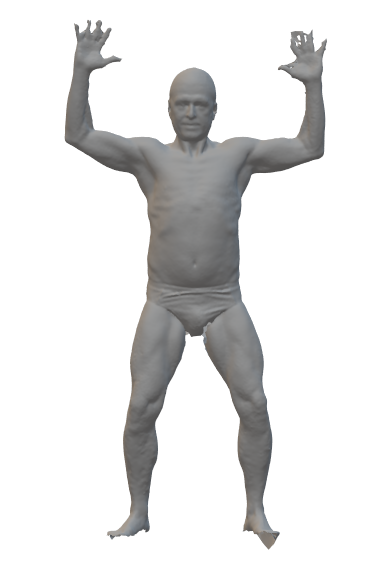}&\includegraphics[height= 0.7 in]{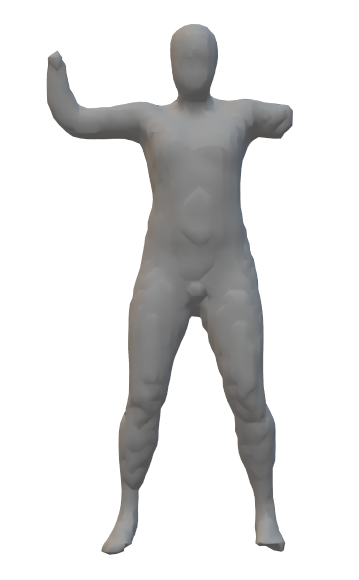}&\includegraphics[height= 0.7 in]{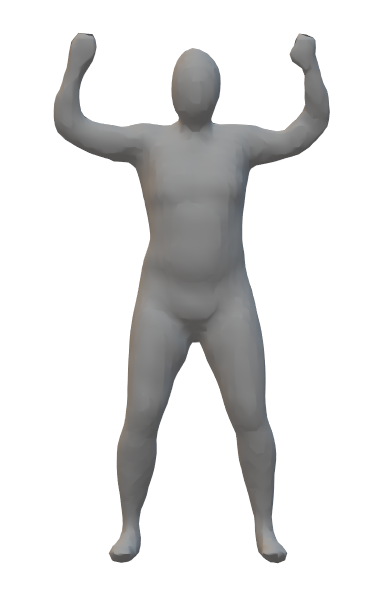}&\includegraphics[height= 0.7 in]{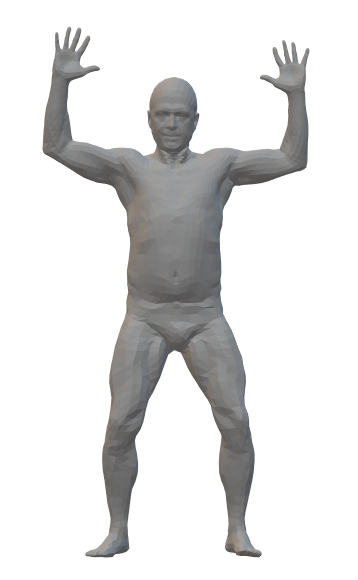}\\
\includegraphics[height= 0.7 in]{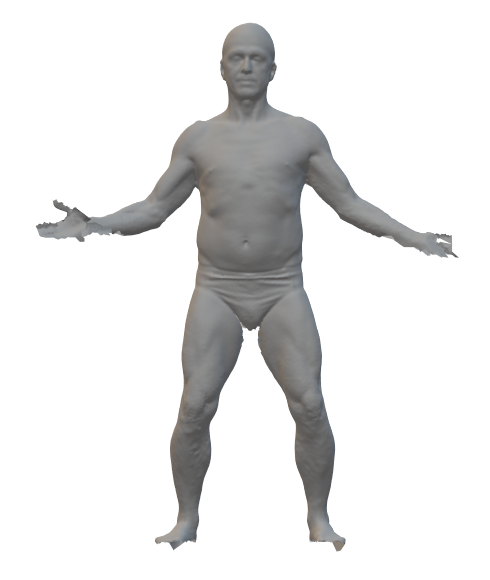}&\includegraphics[height= 0.7 in]{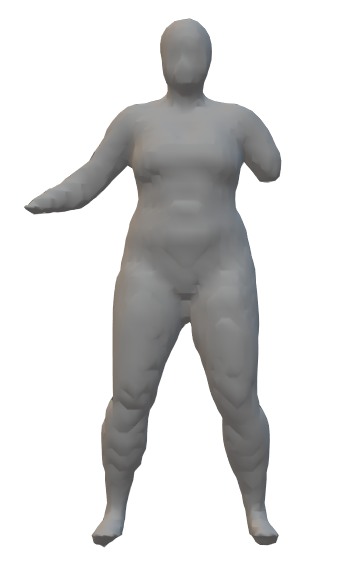}&\includegraphics[height= 0.7 in]{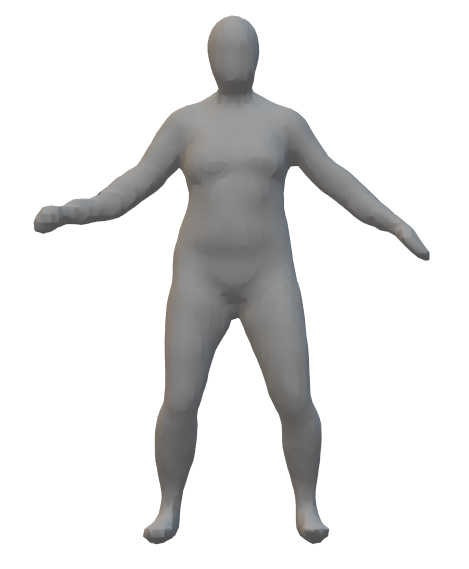}&\includegraphics[height= 0.7 in]{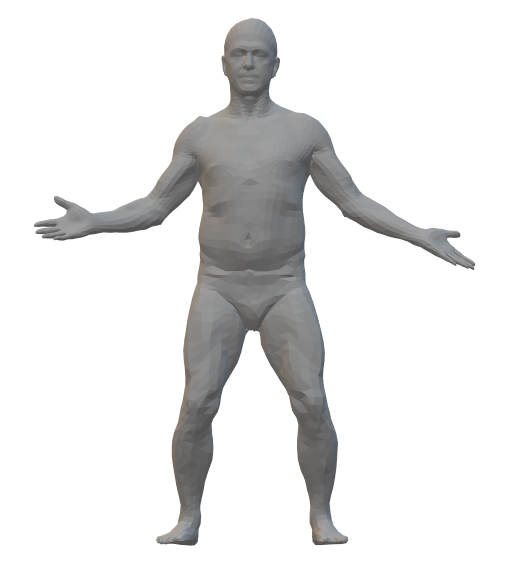}&&\includegraphics[height= 0.7 in]{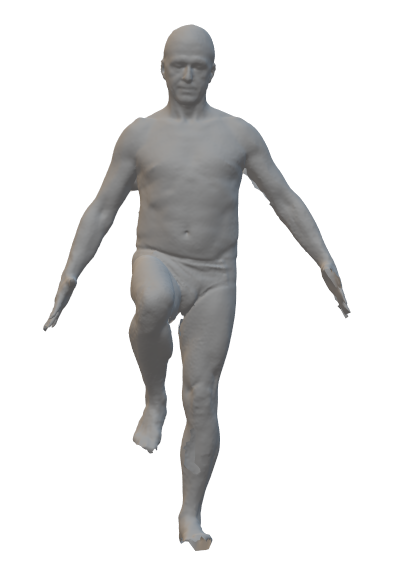}&\includegraphics[height= 0.7 in]{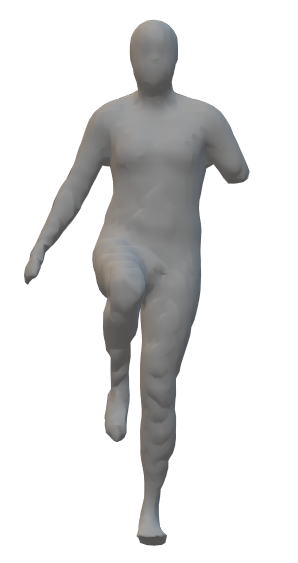}&\includegraphics[height= 0.7 in]{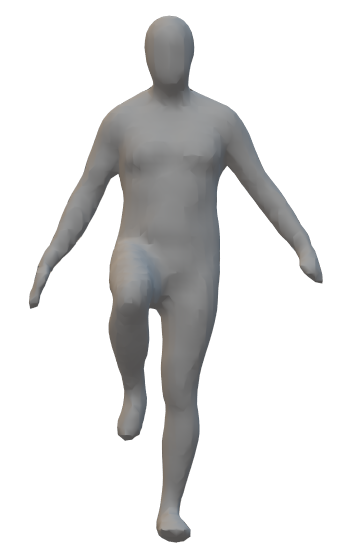}&\includegraphics[height= 0.7 in]{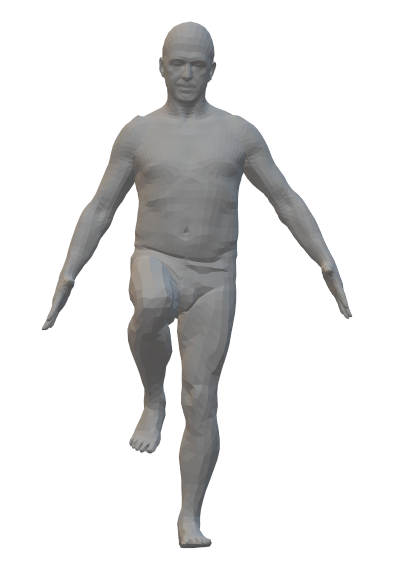}&&\includegraphics[height= 0.7 in]{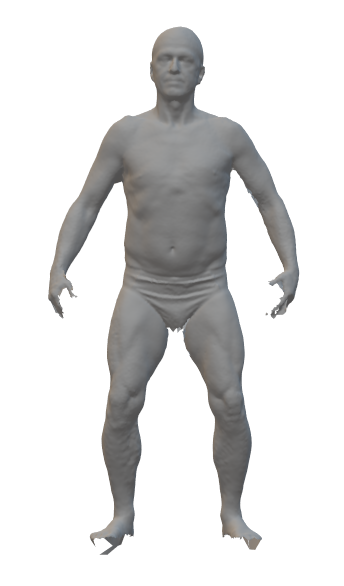}&\includegraphics[height= 0.7 in]{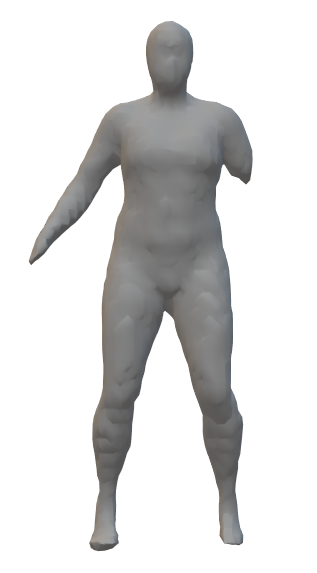}&\includegraphics[height= 0.7 in]{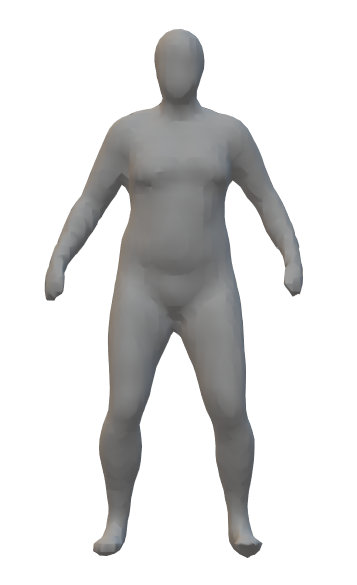}&\includegraphics[height= 0.7 in]{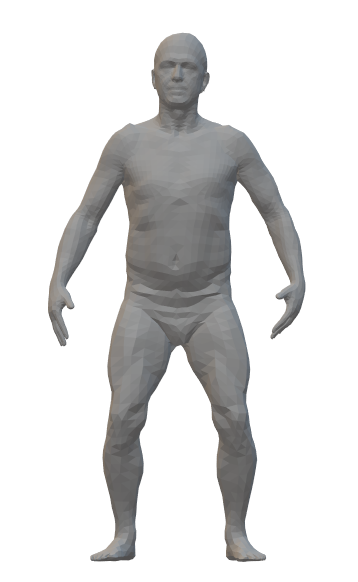}\\
\includegraphics[height= 0.7 in]{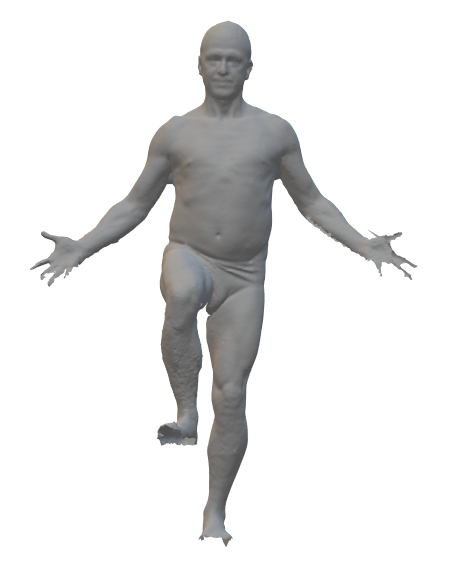}&\includegraphics[height= 0.7 in]{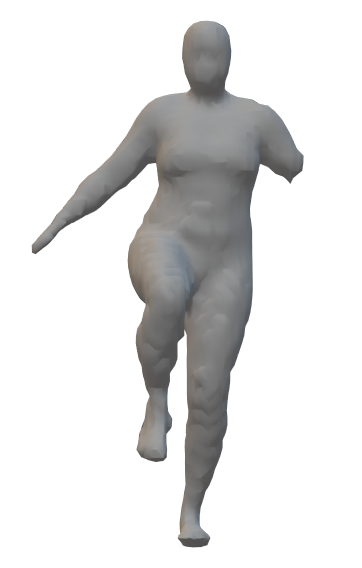}&\includegraphics[height= 0.7 in]{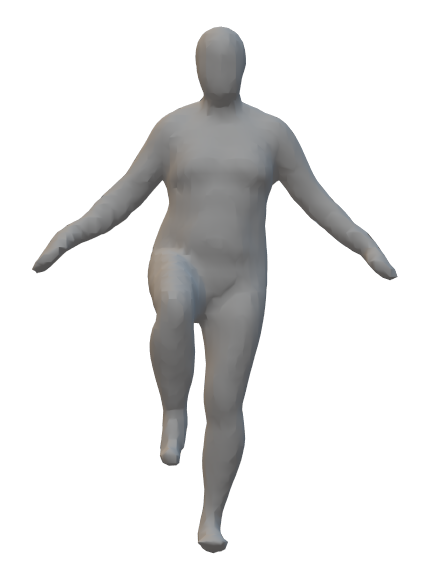}&\includegraphics[height= 0.7 in]{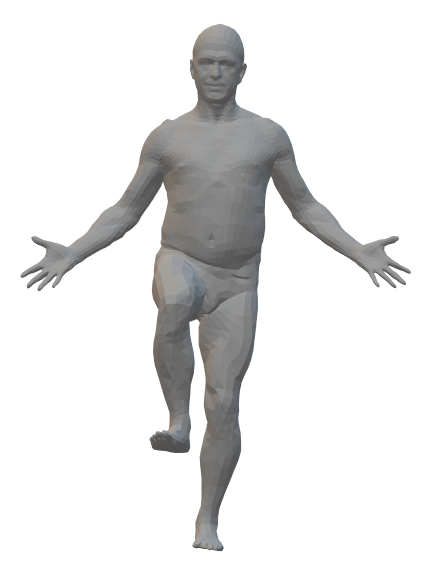}&&\includegraphics[height= 0.7 in]{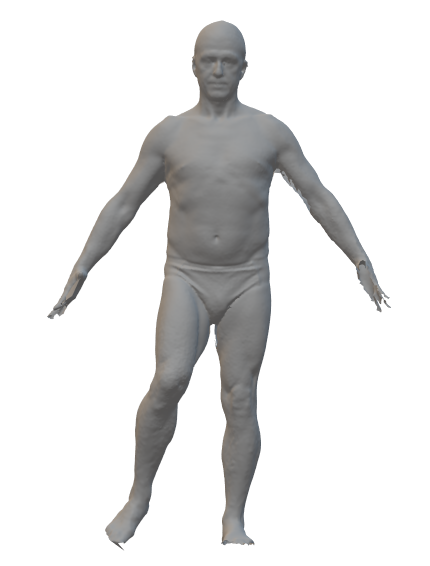}&\includegraphics[height= 0.7 in]{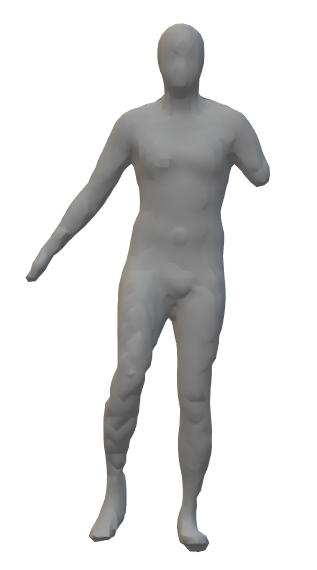}&\includegraphics[height= 0.7 in]{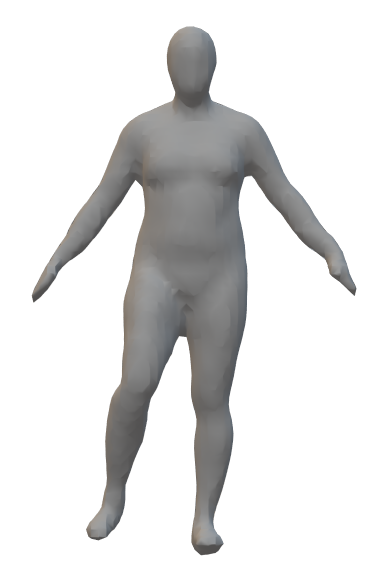}&\includegraphics[height= 0.7 in]{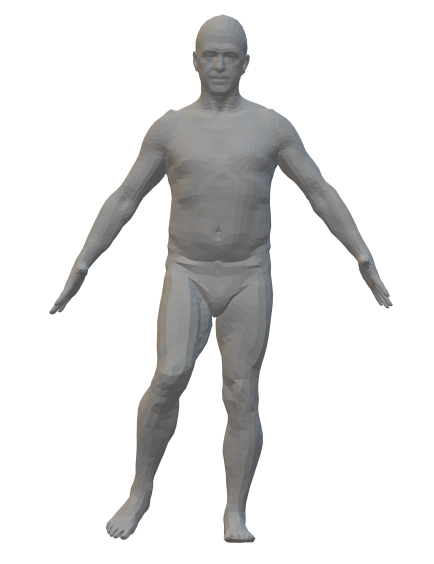}&&\includegraphics[height= 0.7 in]{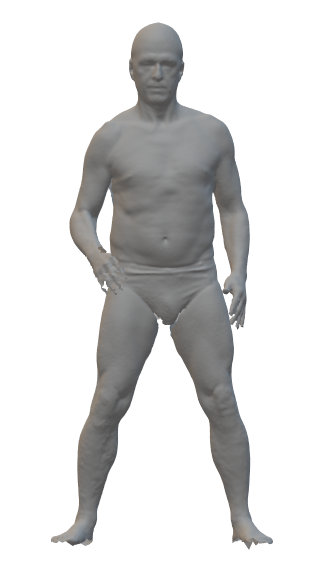}&\includegraphics[height= 0.7 in]{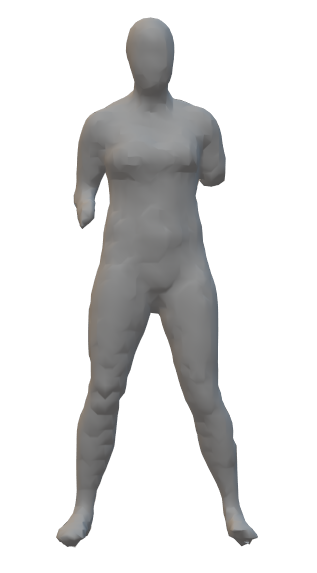}&\includegraphics[height= 0.7 in]{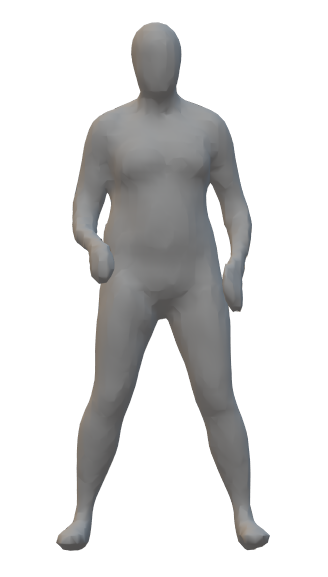}&\includegraphics[height= 0.7 in]{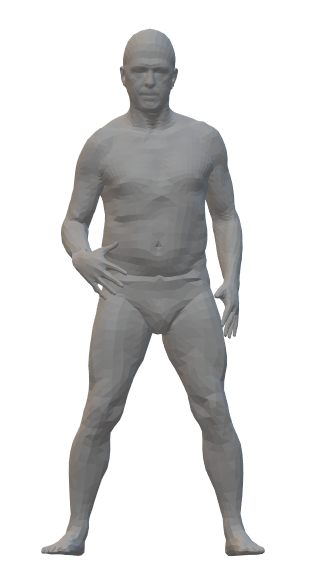}\\
\includegraphics[height= 0.7 in]{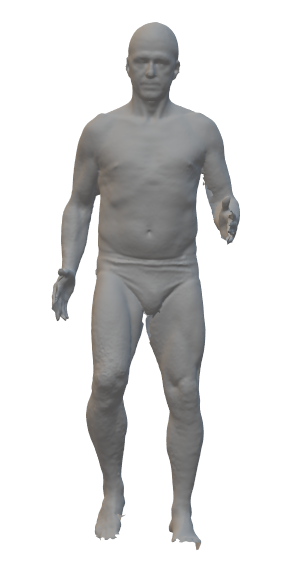}&\includegraphics[height= 0.7 in]{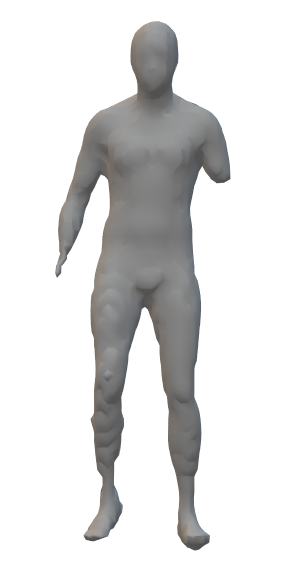}&\includegraphics[height= 0.7 in]{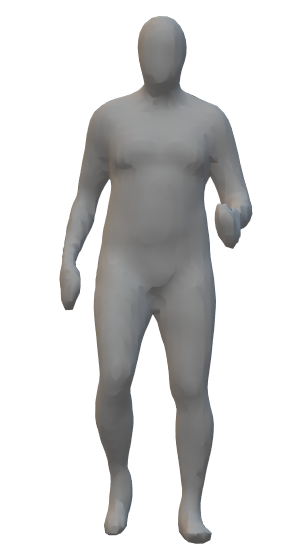}&\includegraphics[height= 0.7 in]{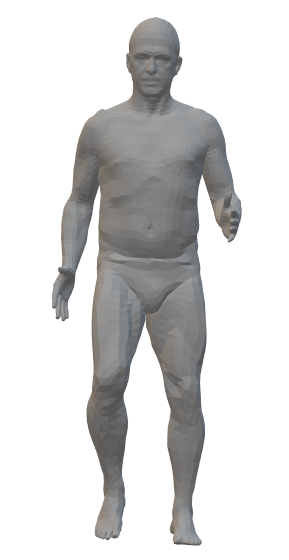}&&\includegraphics[height= 0.7 in]{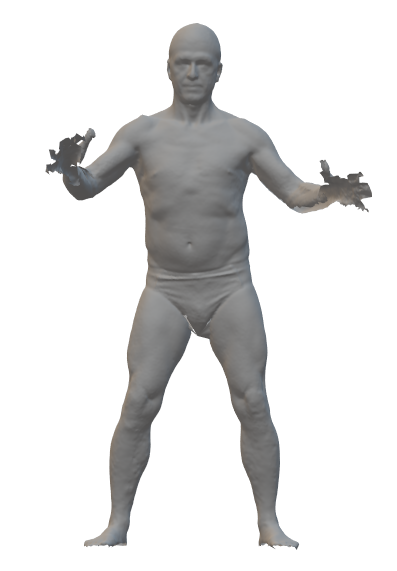}&\includegraphics[height= 0.7 in]{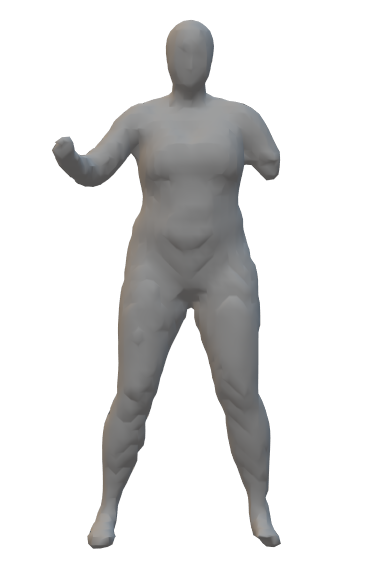}&\includegraphics[height= 0.7 in]{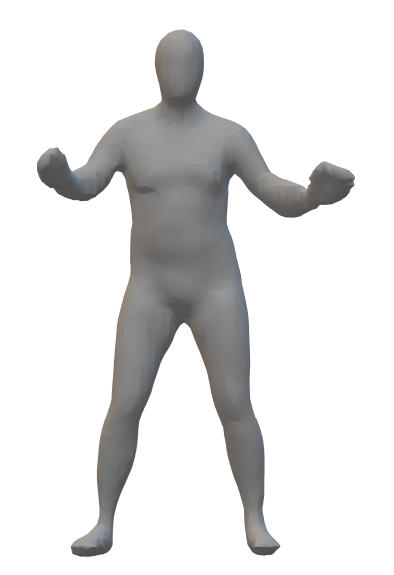}&\includegraphics[height= 0.7 in]{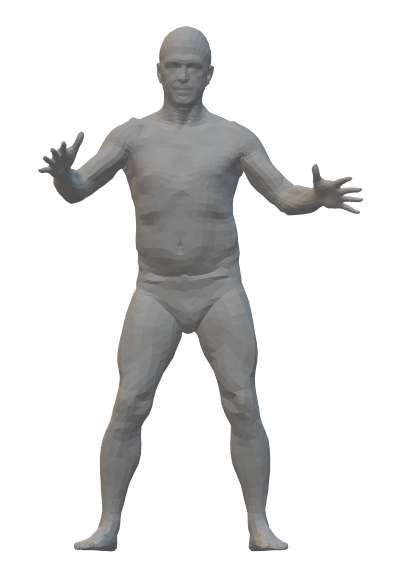}&&\includegraphics[height= 0.7 in]{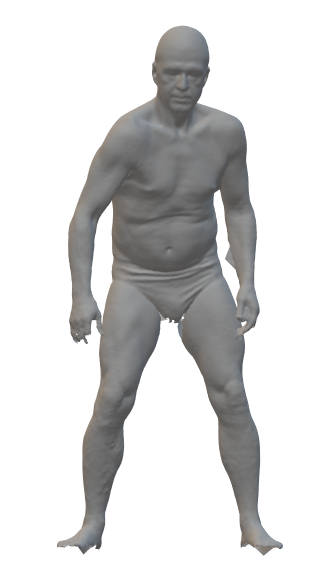}&\includegraphics[height= 0.7 in]{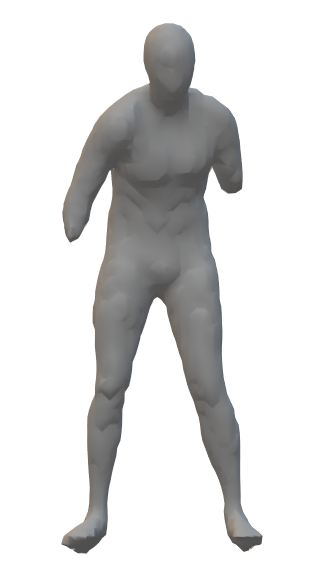}&\includegraphics[height= 0.7 in]{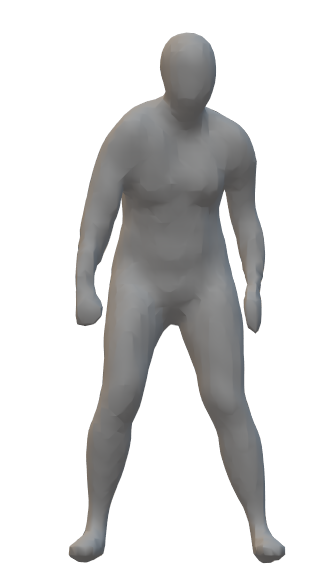}&\includegraphics[height= 0.7 in]{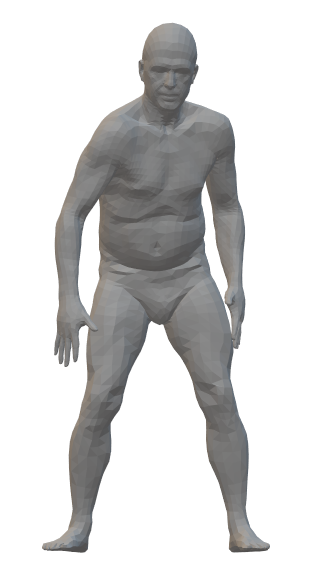}\\
\end{tabular}
\end{adjustbox}
\end{center}
   \caption{The reconstructed unseen human (male) shapes from the D-Faust dataset. Each row consists of three cases with four subfigures: (from left to right) input test scan, 500 epoch-trained baseline architecture, 500-epoch trained LightSAL architecture, ground truth. Noteworthy is the leftmost case from row 2: the baseline architecture reconstructs the input scan into a somewhat different human shape than what the ground truth indicates. The proposed architecture has not been observed to suffer from such behavior.}
\label{reconstructed_human_shape3:Dfaust_dataset_unseenHuman(male)}
\end{figure*}

\begin{figure*}
\begin{center}
\begin{adjustbox}{max size={\textwidth}{\textheight}}
\begin{tabular}{c c c c c c c c c c c c c c}
\includegraphics[height= 0.7 in]{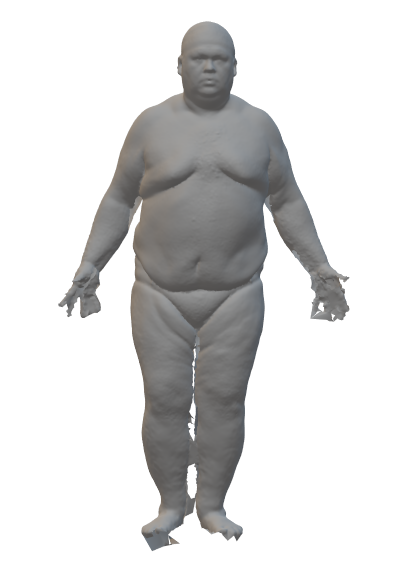}&\includegraphics[ height= 0.7 in]{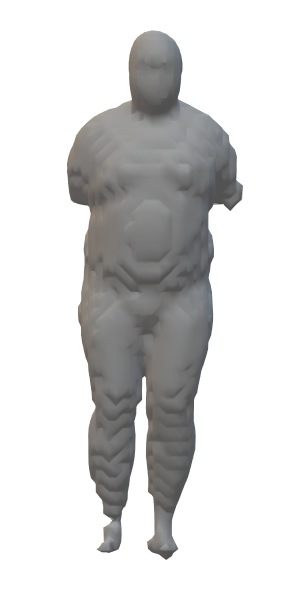}&\includegraphics[ height= 0.7 in]{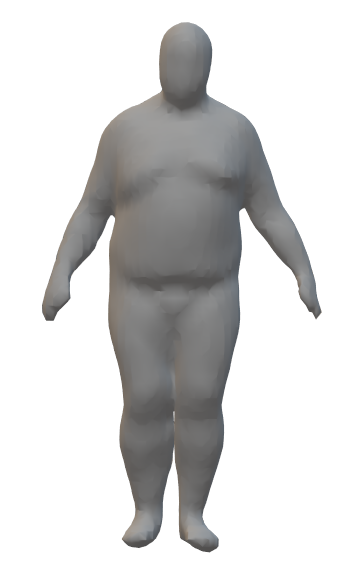}&\includegraphics[ height= 0.7 in]{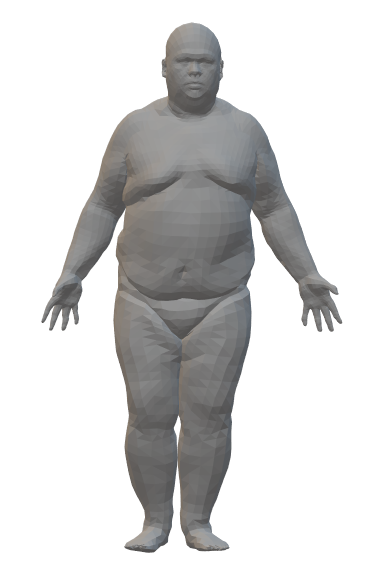}&&\includegraphics[ height= 0.7 in]{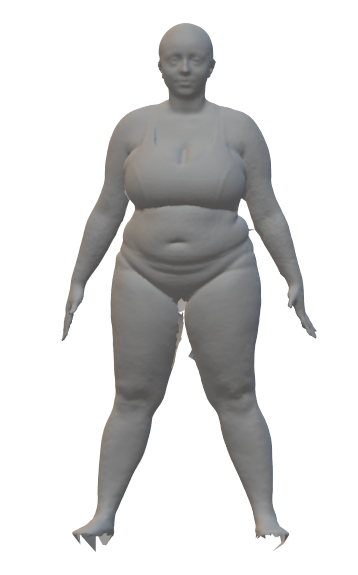}&\includegraphics[ height= 0.7 in]{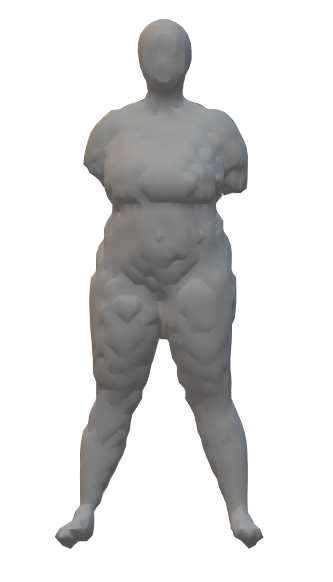}&\includegraphics[ height= 0.7 in]{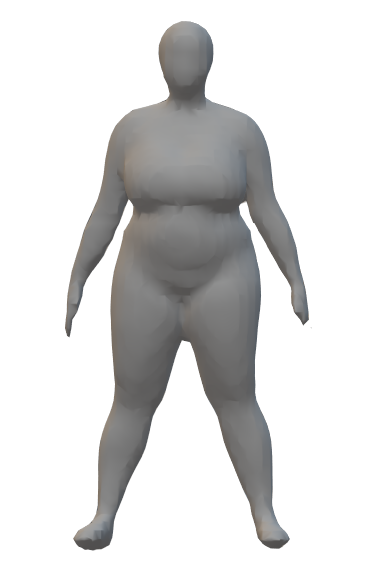}&\includegraphics[ height= 0.7 in]{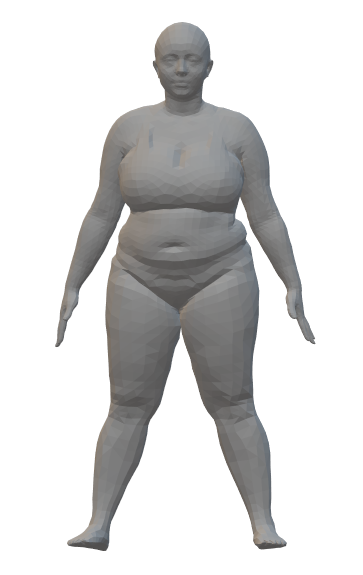}&&\includegraphics[ height= 0.7 in]{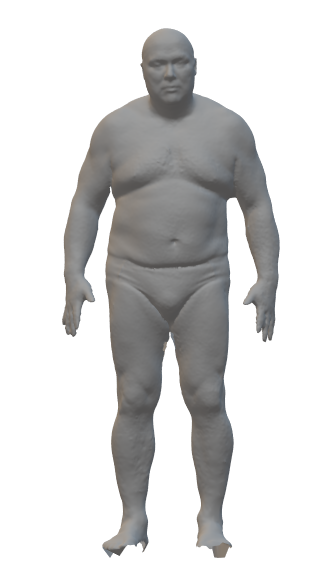}&\includegraphics[ height= 0.7 in]{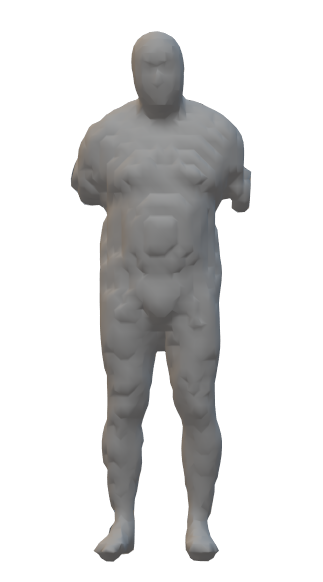}&\includegraphics[ height= 0.7 in]{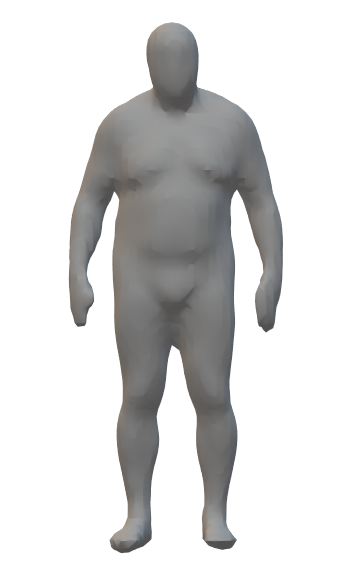}&\includegraphics[ height= 0.7 in]{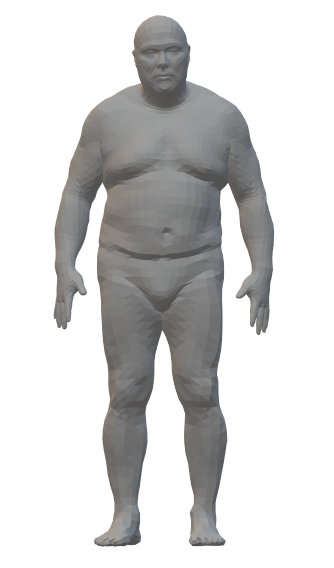}\\
\includegraphics[height= 0.7 in]{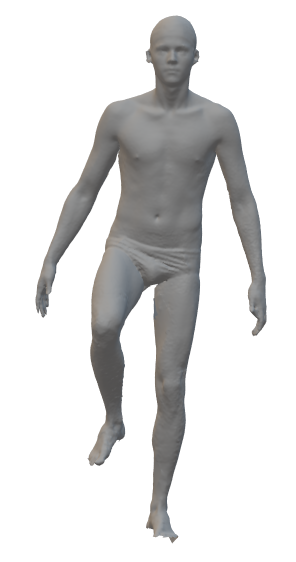}&\includegraphics[ height= 0.7 in]{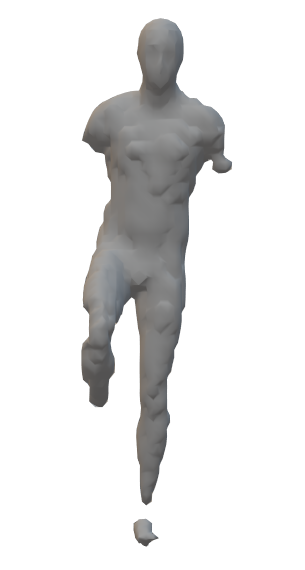}&\includegraphics[ height= 0.7 in]{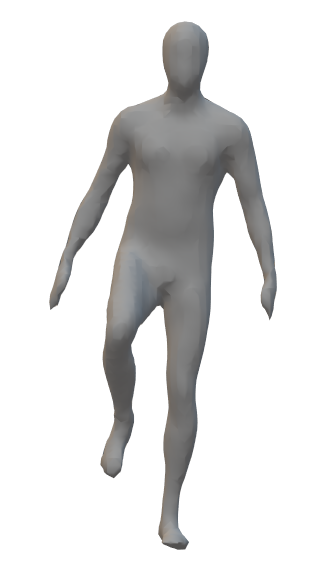}&\includegraphics[ height= 0.7 in]{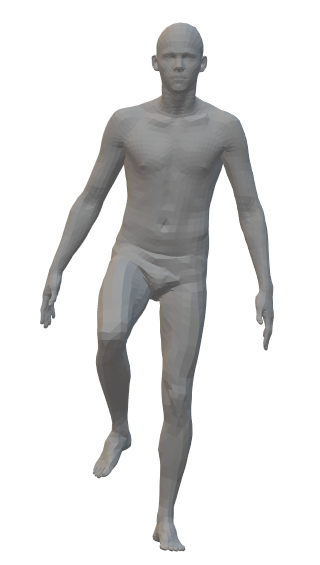}&&\includegraphics[ height= 0.7 in]{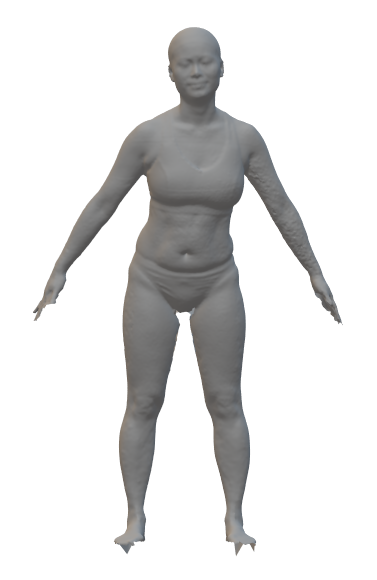}&\includegraphics[height= 0.7 in]{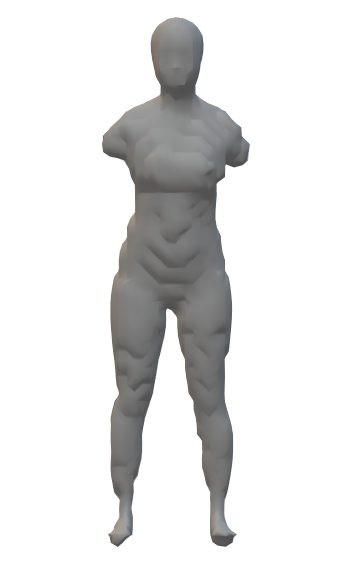}&\includegraphics[height= 0.7 in]{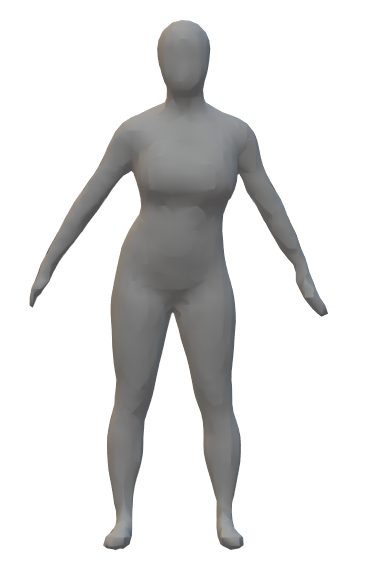}&\includegraphics[height= 0.7 in]{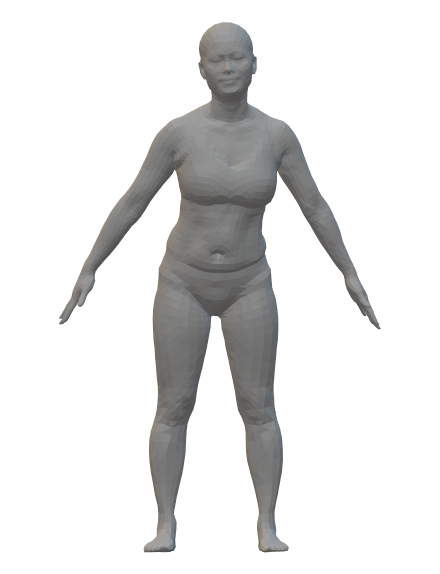}&&\includegraphics[height= 0.7 in]{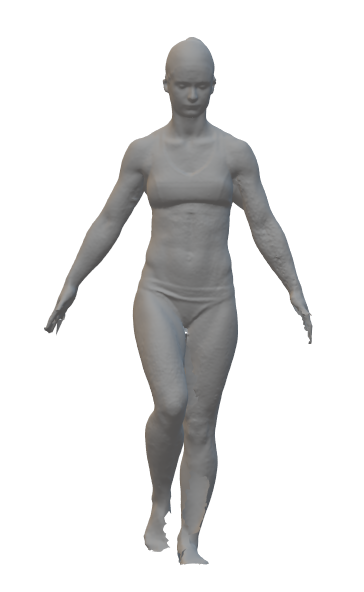}&\includegraphics[height= 0.7 in]{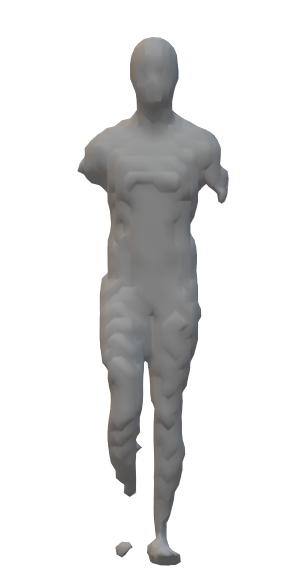}&\includegraphics[height= 0.7 in]{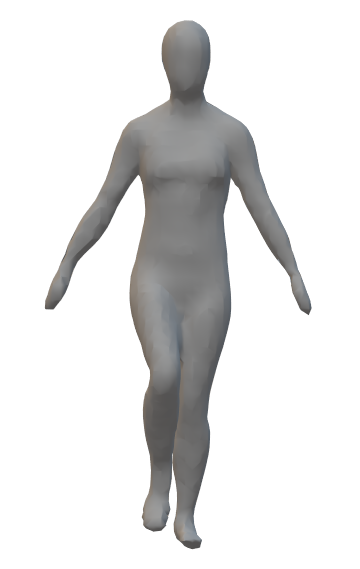}&\includegraphics[height= 0.7 in]{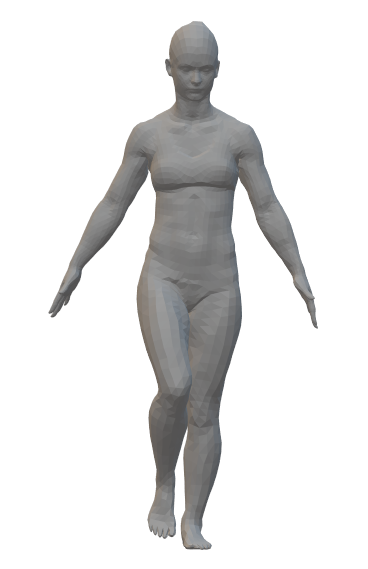}\\
\includegraphics[height= 0.7 in]{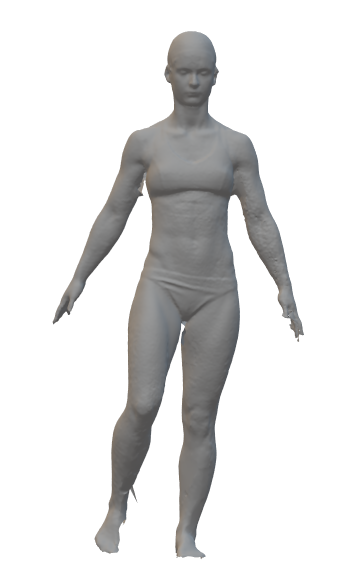}&\includegraphics[height= 0.7 in]{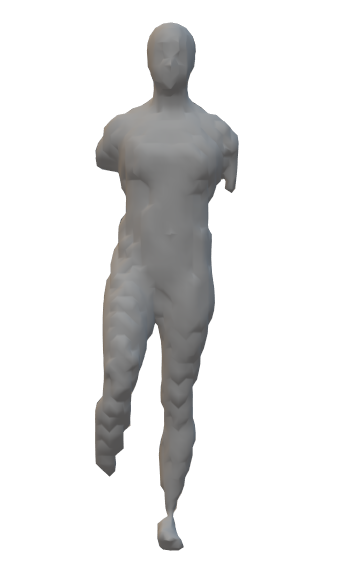}&\includegraphics[height= 0.7 in]{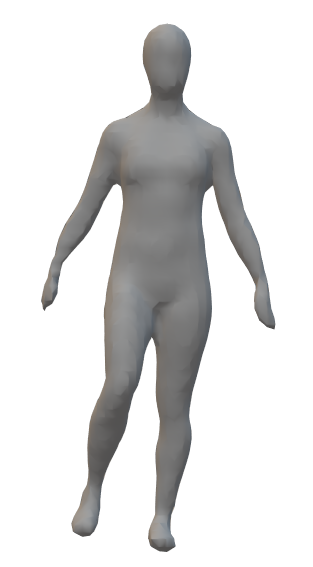}&\includegraphics[height= 0.7 in]{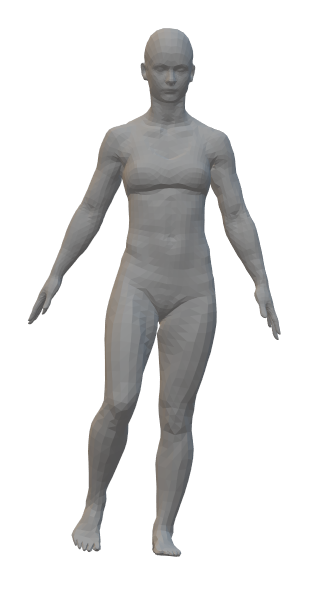}&&\includegraphics[height= 0.7 in]{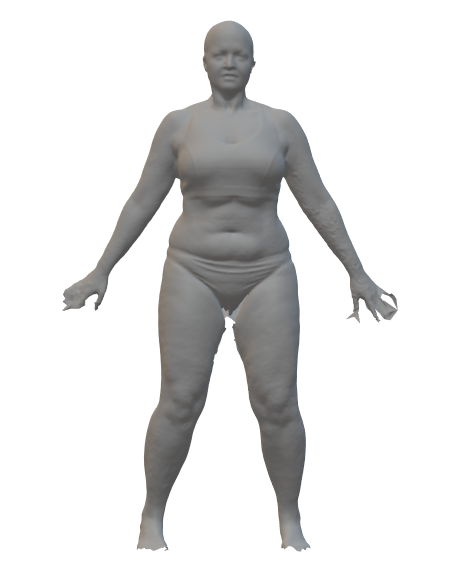}&\includegraphics[ height= 0.7 in]{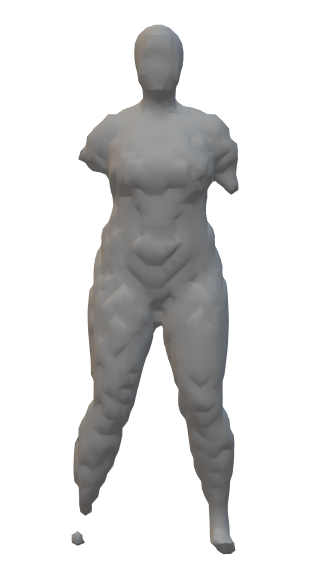}&\includegraphics[height= 0.7 in]{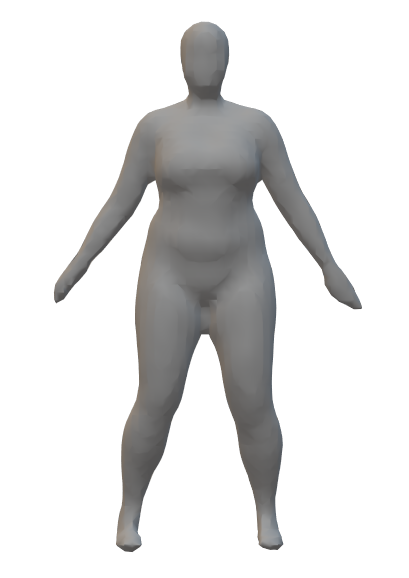}&\includegraphics[height= 0.7 in]{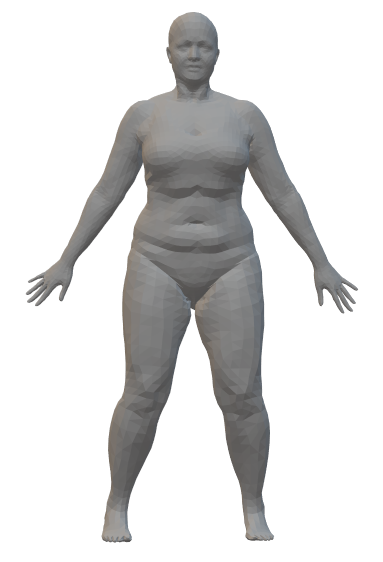}&&\includegraphics[height= 0.7 in]{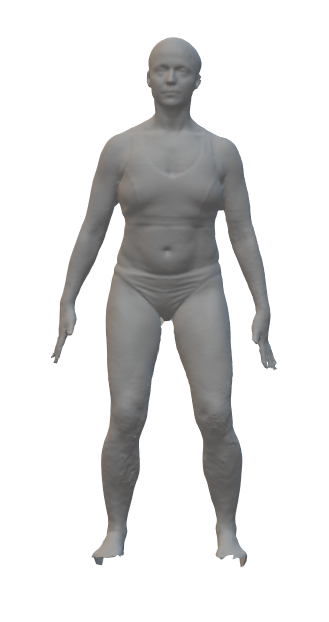}&\includegraphics[height= 0.7 in]{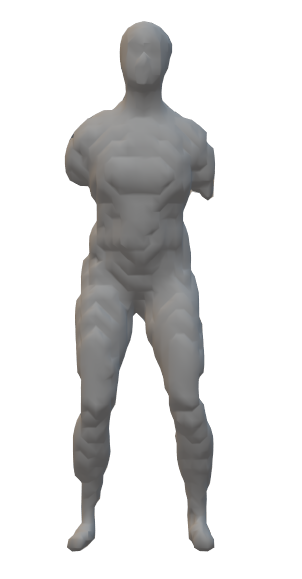}&\includegraphics[height= 0.7 in]{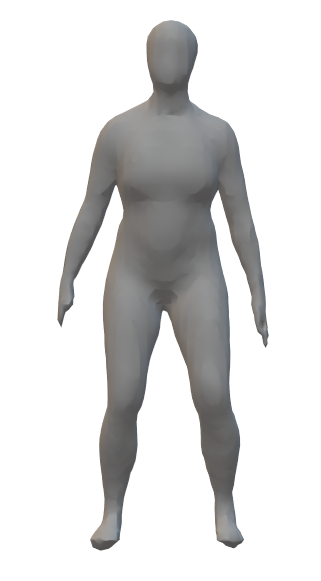}&\includegraphics[height= 0.7 in]{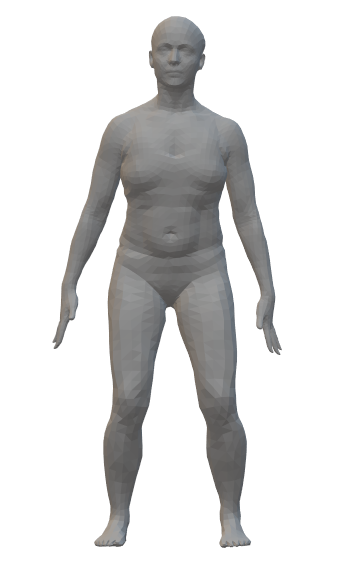}\\
\includegraphics[height= 0.7 in]{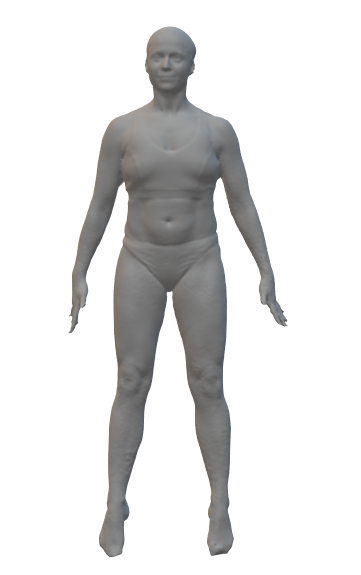}&\includegraphics[height= 0.7 in]{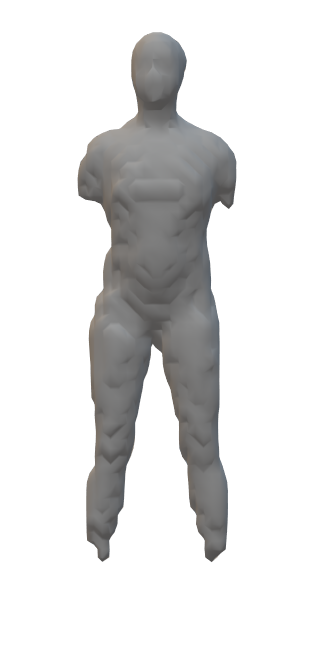}&\includegraphics[height= 0.7 in]{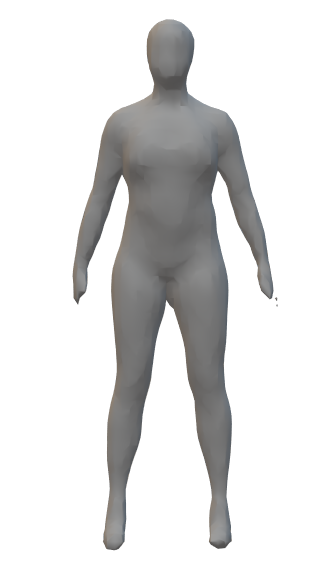}&\includegraphics[height= 0.7 in]{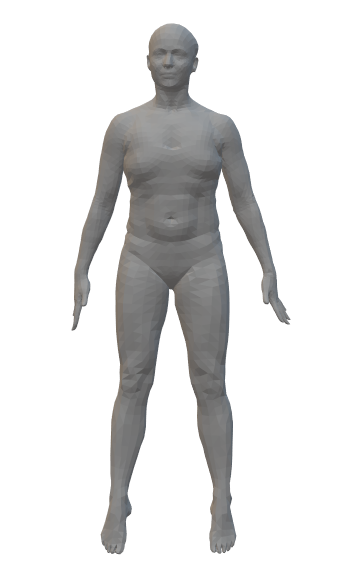}&&\includegraphics[height= 0.7 in]{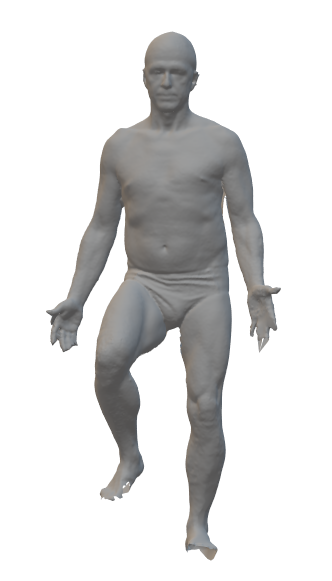}&\includegraphics[height= 0.7 in]{f3_sl_50026_running_on_spot.000128.png}&\includegraphics[ height= 0.7 in]{f3_r_50026_running_on_spot.000128.png}&\includegraphics[height= 0.7 in]{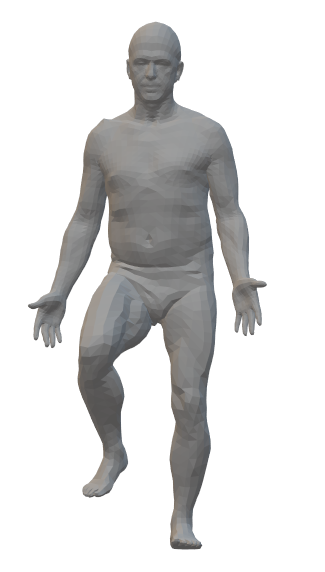}&&\includegraphics[ height= 0.7 in]{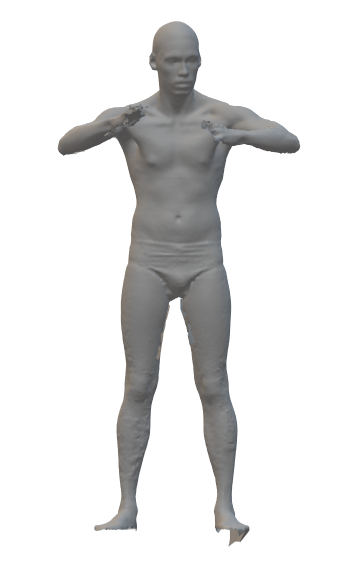}&\includegraphics[ height= 0.7 in]{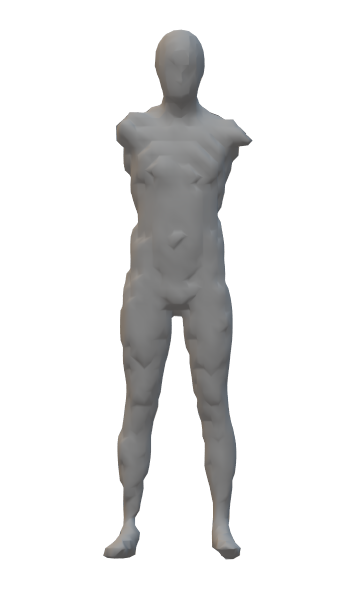}&\includegraphics[ height= 0.7 in]{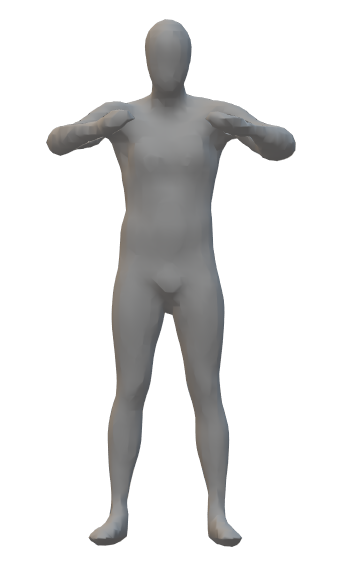}&\includegraphics[ height= 0.7 in]{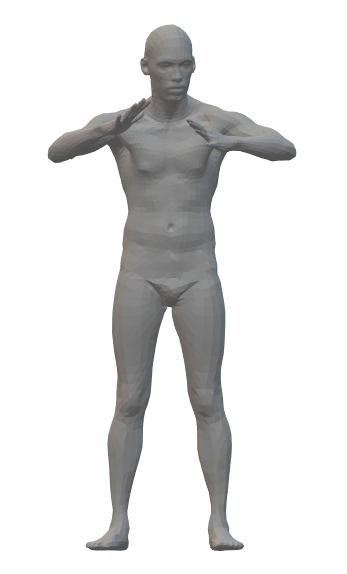}\\
\end{tabular}
\end{adjustbox}
\end{center}
   \caption{Reconstructed unseen poses. Each row consists of three cases with four subfigures: (from left to right) input test scan, baseline SAL trained with 500 epochs, LightSAL trained with 500 epochs, ground truth.}
\label{reconstructed_human_shape3:Dfaust_dataset_unseenPose}
\end{figure*}

For further verification, and to compensate for the unavailable unseen shape generation models, 
we conducted an additional experiment, with visual results reported in Figures~\ref{reconstructed_human_shape2:Dfaust_dataset_unseenHuman_female}, \ref{reconstructed_human_shape3:Dfaust_dataset_unseenHuman(male)}, and \ref{reconstructed_human_shape3:Dfaust_dataset_unseenPose}, whereas the corresponding numerical results are in Table~\ref{ablation_study:Chamfer_distances}. In this setting, both the baseline and the LightSAL architectures were trained with 500 epochs. The results show that whereas the baseline model clearly has not converged yet, the LightSAL model has already achieved high reconstruction quality.

Most interestingly, baseline SAL indeed suffers from unwanted behavior that was already reported in Figure 7 of the SAL publication \cite{Atzmon_2020_CVPR}: in some cases baseline SAL starts to reconstruct a totally different human shape than what the ground truth and the input scan indicate. Our experiments confirm this behavior for baseline SAL (See Figure~\ref{reconstructed_human_shape3:Dfaust_dataset_unseenHuman(male)}). \textit{In contrast, LightSAL has not been observed to suffer from such unwanted behavior.}

As a final note, \cite{Atzmon_2020_CVPR} also provides numbers from using latent optimization for improved quality; this option was not adopted for our experiments, as the implementation of the latent optimization procedure was not clear based on \cite{Atzmon_2020_CVPR}. 

\begin{table}
\begin{center}
\begin{adjustbox}{width=\columnwidth}
\begin{tabular}{|c| c|c|c|c|c|c|c|} 
\multicolumn{8}{c}{} \\ \cline{1-8}
\multirow{2}{*}{Type}&\multirow{2}{*}{Method}&\multicolumn{3}{c}{Registrations}&\multicolumn{3}{|c|}{Scans} \\ \cline{3-8}
&&5\%&50\%&95\%&5\%&50\%&95\% \\ \cline{1-8}
\multicolumn{1}{|c|}{}&SAL \cite{Atzmon_2020_CVPR} (reconstruction)&1.18&3.08&11.41&0.06&0.10&0.18 \\ \cline{2-8}
\multicolumn{1}{|c}{}&\multicolumn{1}{|c|}{Ours (reconstruction)}&\textbf{0.09}&\textbf{0.15}&\textbf{0.42}&0.06&\textbf{0.09}&\textbf{0.16}\\ \cline{2-8}
\multicolumn{1}{|c}{}&\multicolumn{1}{|c|}{SAL \cite{Atzmon_2020_CVPR} (unseen human)}&1.06&2.66&9.42&0.16&0.40&1.40\\ \cline{2-8}
\multicolumn{1}{|c}{Test}&\multicolumn{1}{|c|}{Ours (unseen human)}&\textbf{0.16}&\textbf{0.34}&\textbf{3.15}&\textbf{0.09}&\textbf{0.17}&\textbf{0.71}\\ \cline{2-8}
\multicolumn{1}{|c}{}&\multicolumn{1}{|c|}{SAL \cite{Atzmon_2020_CVPR} (unseen pose)}&1.76&4.96&17.06&0.09&0.19&0.92\\ \cline{2-8}
\multicolumn{1}{|c}{}&\multicolumn{1}{|c|}{Ours (unseen pose)}&\textbf{0.09}&\textbf{0.19}&\textbf{1.06}&\textbf{0.06}&\textbf{0.11}&\textbf{0.31}\\ \cline{1-8}
\multicolumn{8}{c}{}
\end{tabular}
\end{adjustbox}
\caption{Performance of baseline SAL and LightSAL architectures when both have been trained with 500 epochs: Chamfer distances of the reconstructed shape against ground truth registrations and raw scans. The data are presented in percentiles ($5{^{th}}$, $50{^{th}}$, and  $95{^{th}}$), values have been multiplied by $10{^3}$.}
\label{ablation_study:Chamfer_distances}
\end{center}
\end{table}

\begin{table}[]
    \centering
    \begin{adjustbox}{width=\columnwidth}
    \begin{tabular}{|c|c|c|c|}
    \hline
       Method  & Training time&Type&\# Trainable parameters  \\ \cline{1-4}
        \multirow{2}{*}{SAL \cite{Atzmon_2020_CVPR}} &111 $\pm$ 3 s&Encoder&2'365'952\\ \cline{3-4}
        &per epoch&Decoder&1'843'195\\ \cline{1-4}
         Ours&65 $\pm$ 3 s&Encoder&658'944\\ \cline{3-4}
         &per epoch&Decoder&363'643\\ \cline{1-4}
         \multicolumn{4}{c}{}\\
    \end{tabular}
    \end{adjustbox}
    \caption{Comparison of training time and number of model parameters between the proposed and baseline architectures.}
    \label{ablation_study:time_parameters}
\end{table}
\subsection{Training time}
To evaluate the training time benefit of LightSAL over baseline SAL, benchmarking was conducted on the workstation used for experiments. Our workstation ran Ubuntu Linux 20.04 LTS, had 256GB of RAM, and was equipped with a 24GB GeForce RTX 3090 GPU. The CPU was Intel Cascade Lake X (Core i9-10900X,
3.70 GHz). The deep learning framework used was PyTorch 1.8.0. The time measurement values in Table~\ref{ablation_study:time_parameters} show that LightSAL requires around 40\% less time per epoch in training. On the other hand, as numerical quality results (Tables \ref{chamfer_distance:Dfaust_Dataset_primary}, \ref{chamfer_distance:Dfaust_Dataset_unseenHuman}, and \ref{chamfer_distance:Dfaust_Dataset_unseenPose}) indicate that equivalent reconstruction quality can be achieved with 75\% less training iterations, LightSAL can be approximated to reduce training time by a factor of 6$\times$ compared to baseline SAL. 

In terms of model size, Table~\ref{ablation_study:time_parameters} shows that the overall model size of the LightSAL encoder-decoder is 75\% smaller than the one of baseline SAL.

\begin{figure*}
\begin{center}
\begin{adjustbox}{max size={\textwidth}{\textheight}}
\begin{tabular}{c c c c c c c c c c c c c c}
\includegraphics[height= 0.7 in]{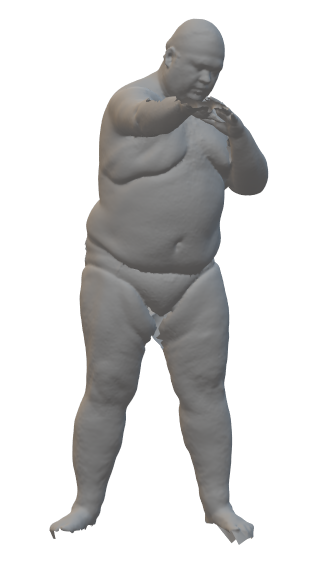}&\includegraphics[ height= 0.7 in]{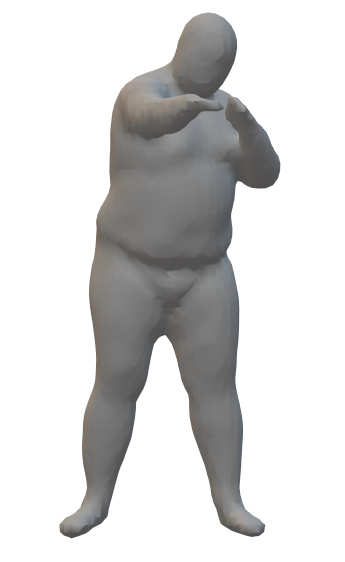}&\includegraphics[ height= 0.7 in]{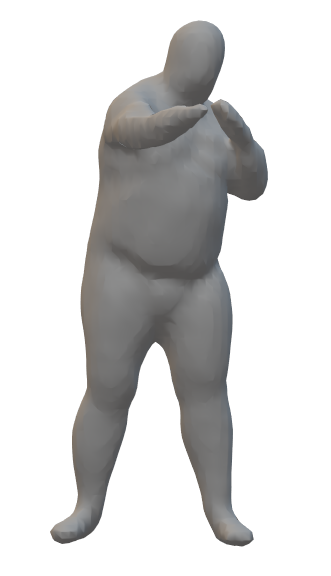}&\includegraphics[ height= 0.7 in]{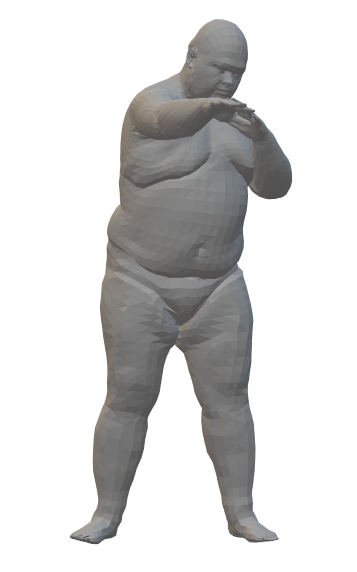}&&\includegraphics[ height= 0.7 in]{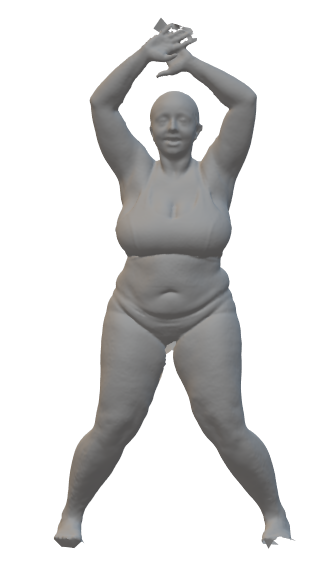}&\includegraphics[ height= 0.7 in]{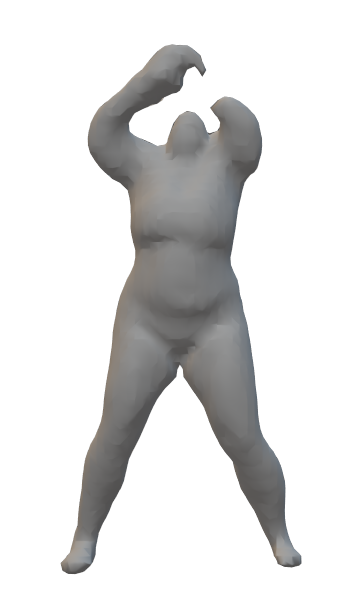}&\includegraphics[ height= 0.7 in]{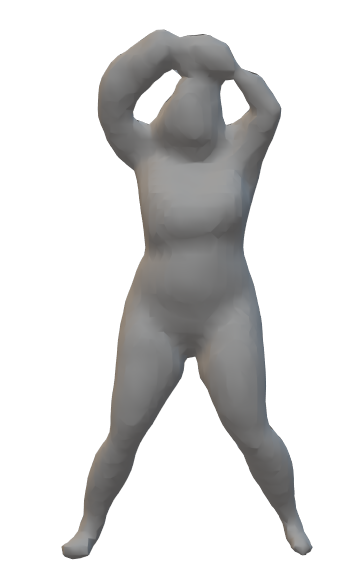}&\includegraphics[ height= 0.7 in]{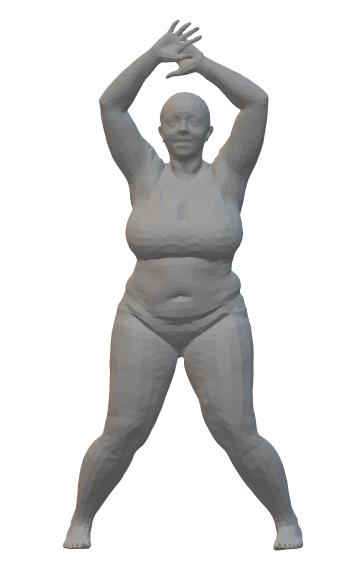}&&\includegraphics[ height= 0.7 in]{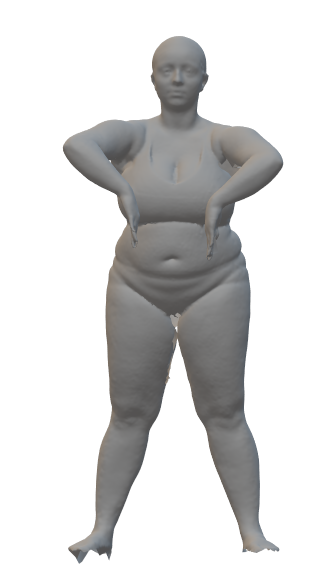}&\includegraphics[ height= 0.7 in]{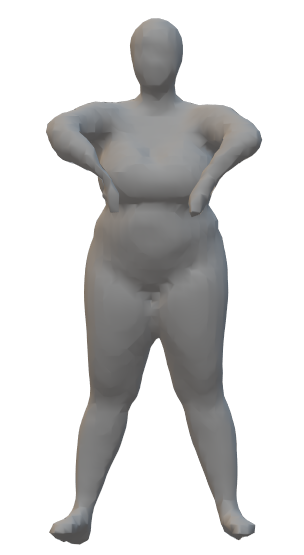}&\includegraphics[ height= 0.7 in]{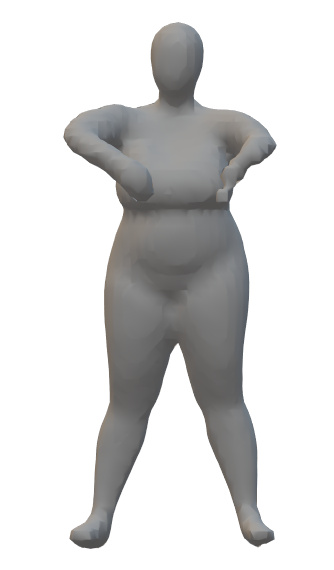}&\includegraphics[ height= 0.7 in]{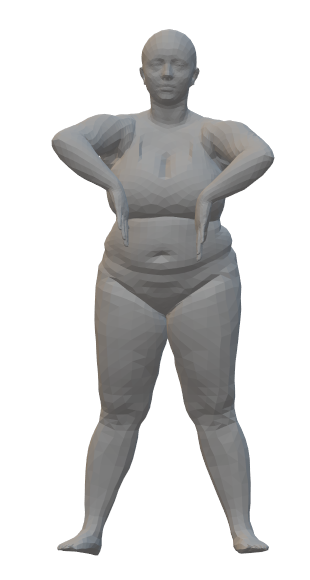}\\
\includegraphics[ height= 0.7 in]{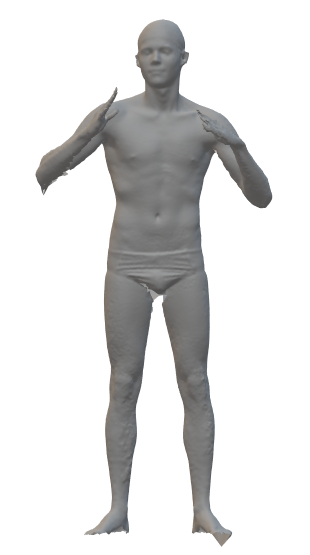}&\includegraphics[ height= 0.7 in]{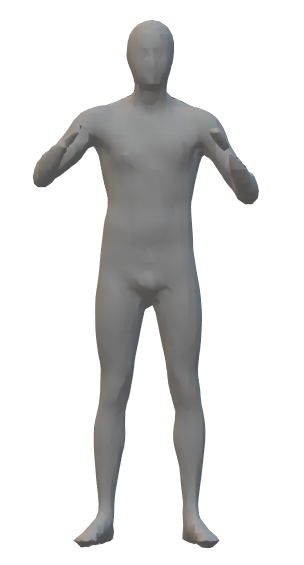}&\includegraphics[ height= 0.7 in]{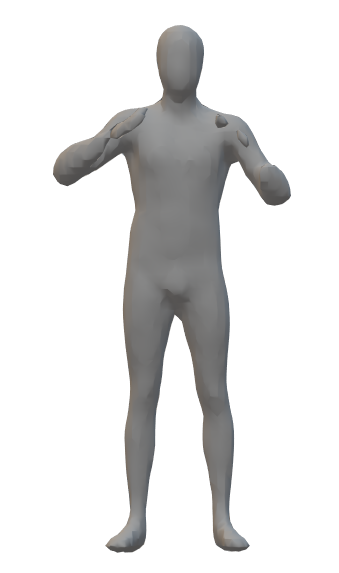}&\includegraphics[ height= 0.7 in]{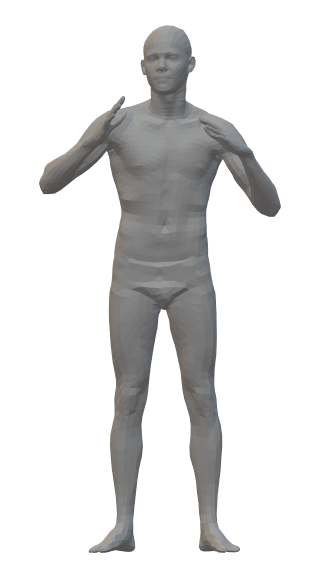}&&\includegraphics[ height= 0.7 in]{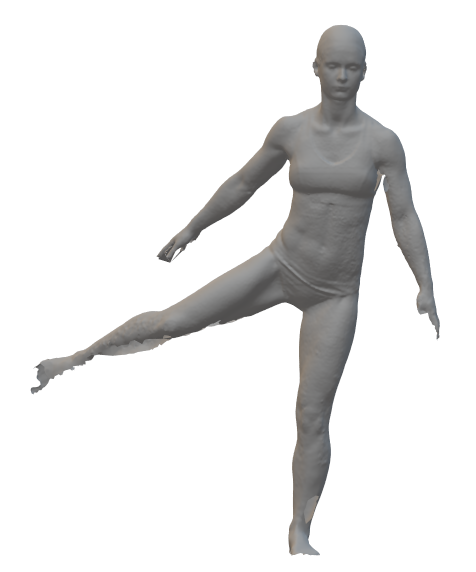}&\includegraphics[ height= 0.7 in]{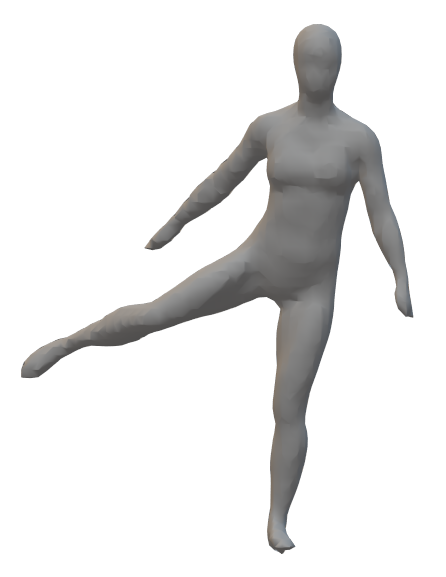}&\includegraphics[ height= 0.7 in]{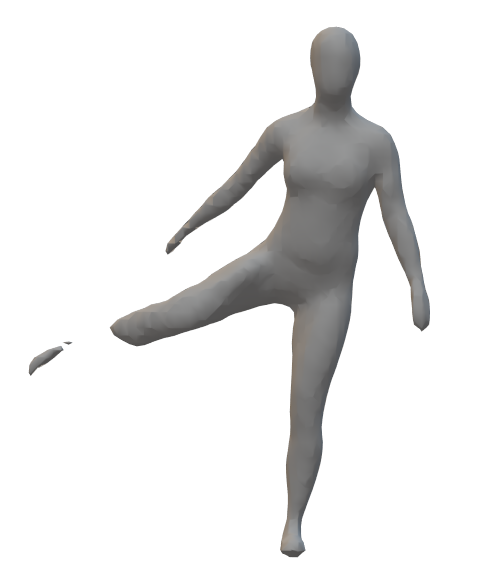}&\includegraphics[ height= 0.7 in]{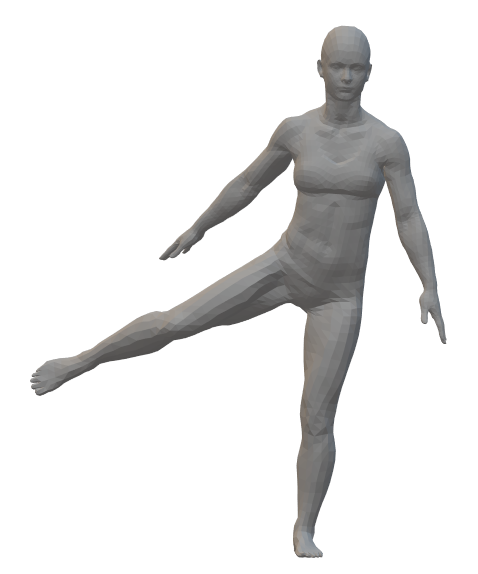}&&\includegraphics[height= 0.7 in]{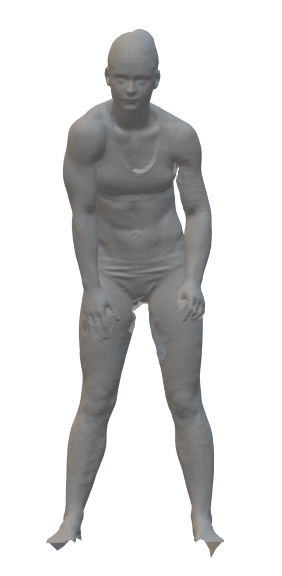}&\includegraphics[height= 0.7 in]{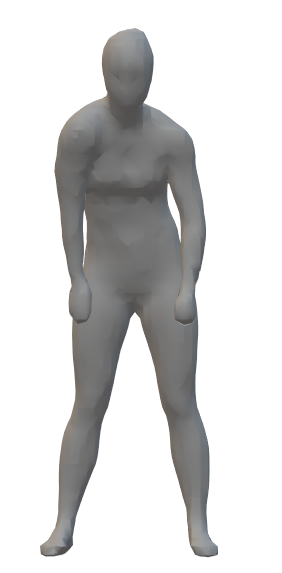}&\includegraphics[height= 0.7 in]{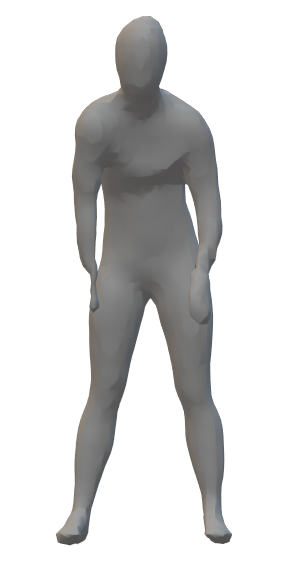}&\includegraphics[height= 0.7 in]{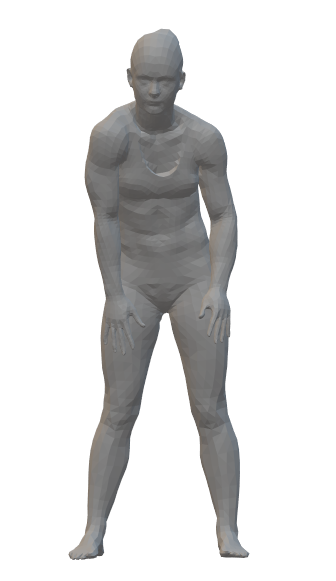}\\
\includegraphics[height= 0.7 in]{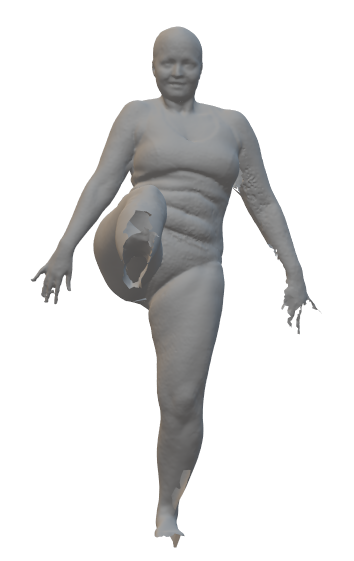}&\includegraphics[height= 0.7 in]{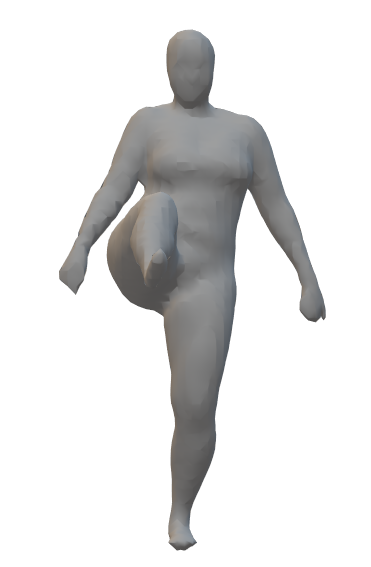}&\includegraphics[height= 0.7 in]{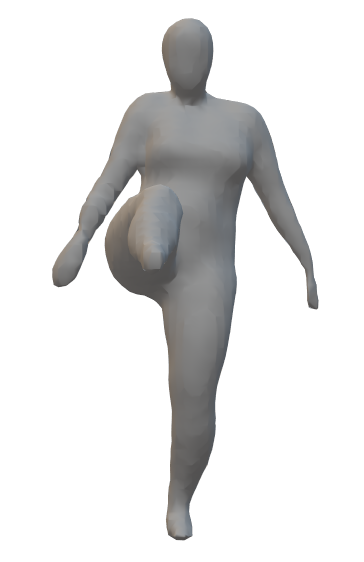}&\includegraphics[height= 0.7 in]{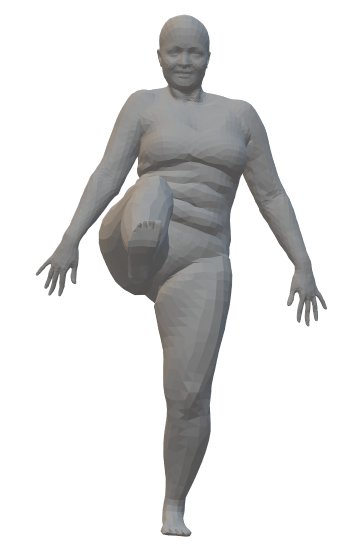}&&\includegraphics[height= 0.7 in]{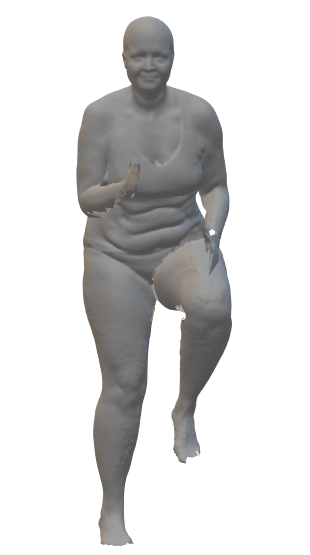}&\includegraphics[ height= 0.7 in]{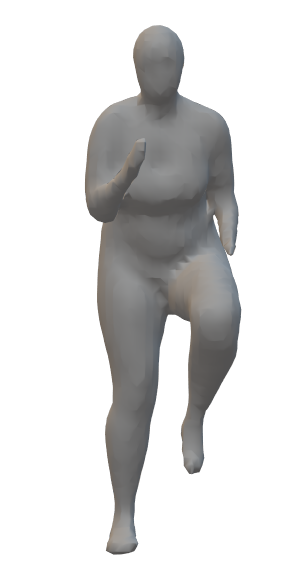}&\includegraphics[ height= 0.7 in]{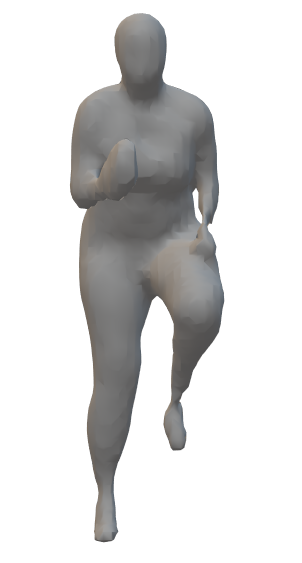}&\includegraphics[height= 0.7 in]{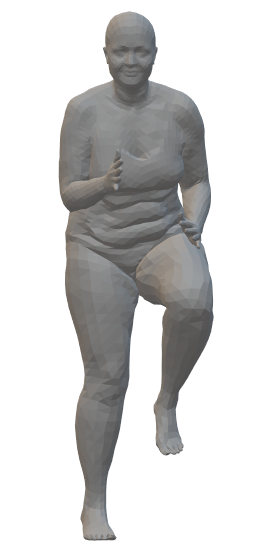}&&\includegraphics[height= 0.7 in]{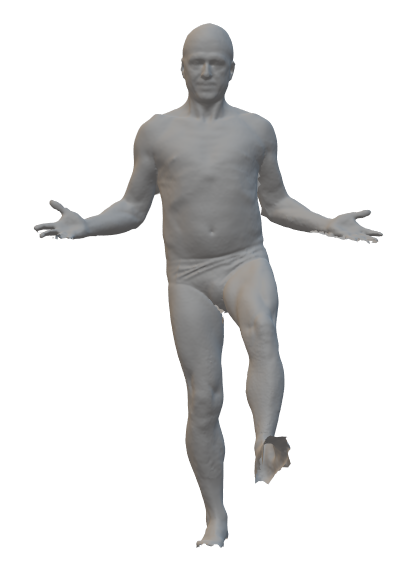}&\includegraphics[height= 0.7 in]{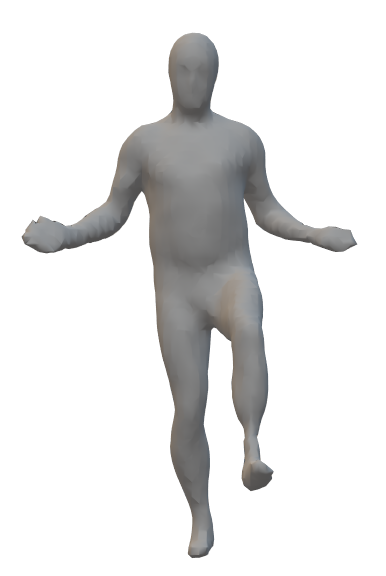}&\includegraphics[height= 0.7 in]{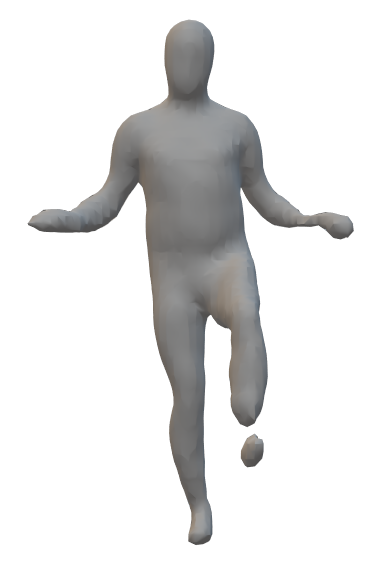}&\includegraphics[ height= 0.7 in]{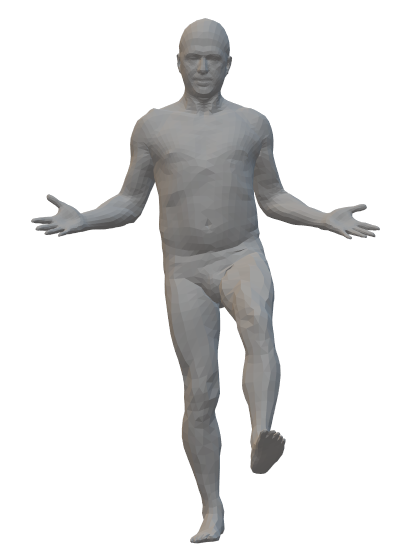}\\
\includegraphics[height= 0.7 in]{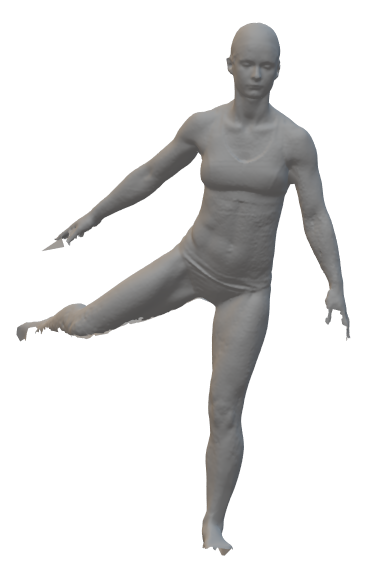}&\includegraphics[height= 0.7 in]{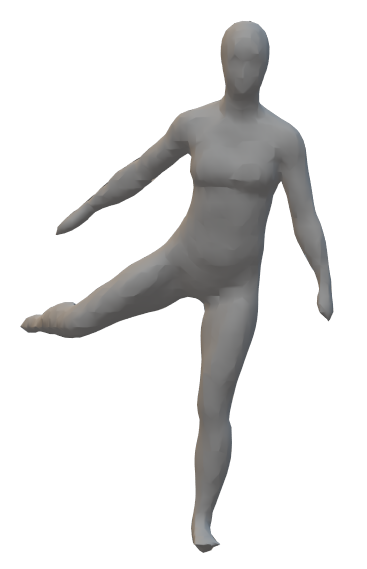}&\includegraphics[height= 0.7 in]{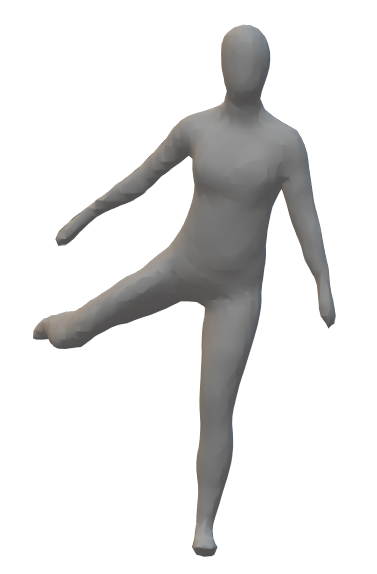}&\includegraphics[height= 0.7 in]{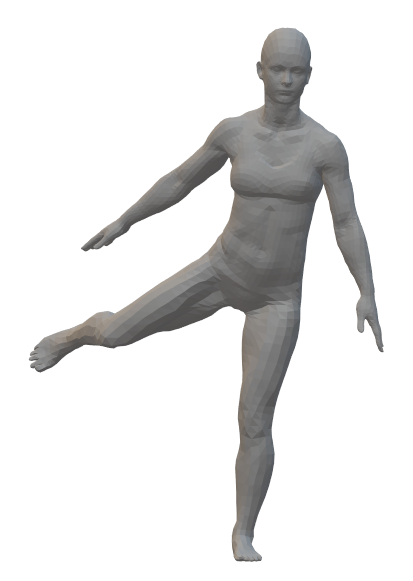}&&\includegraphics[height= 0.7 in]{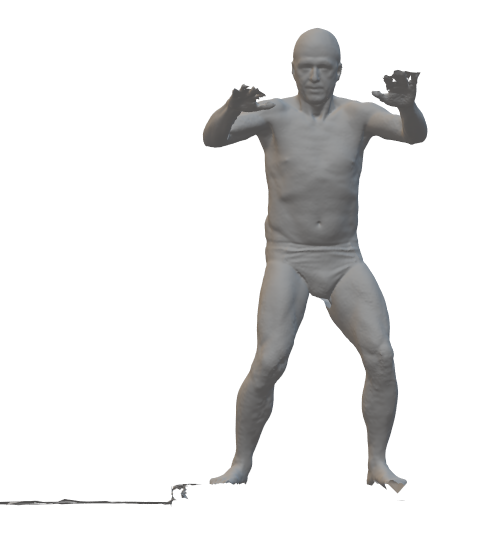}&\includegraphics[height= 0.7 in]{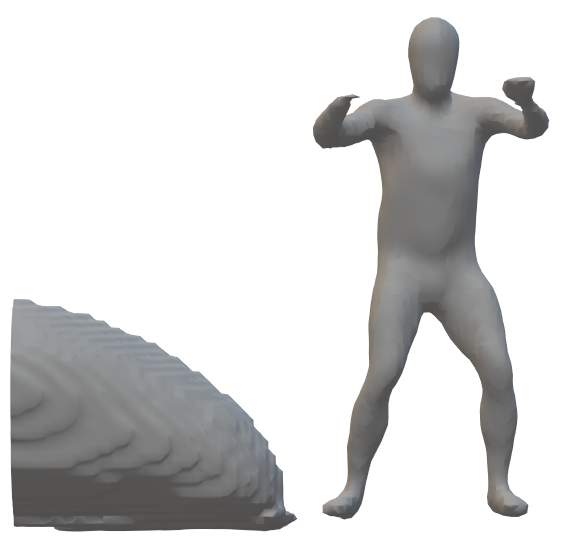}&\includegraphics[ height= 0.7 in]{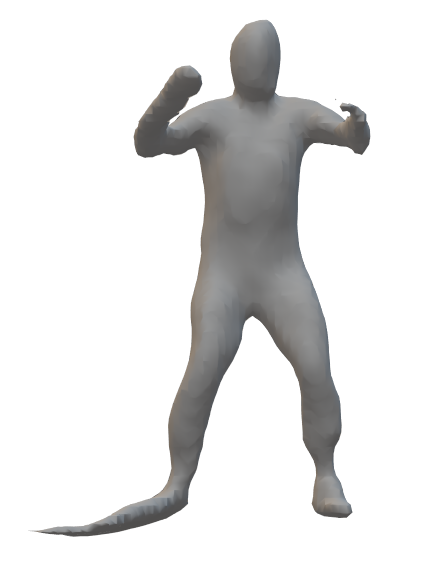}&\includegraphics[ height= 0.7 in]{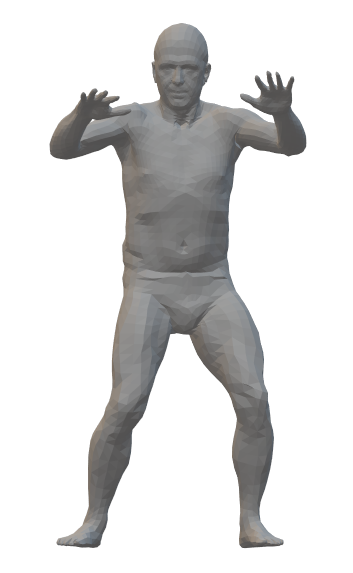}&&\includegraphics[ height= 0.7 in]{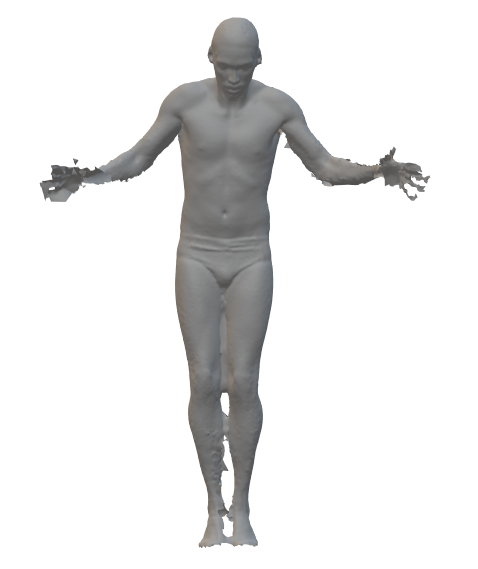}&\includegraphics[ height= 0.7 in]{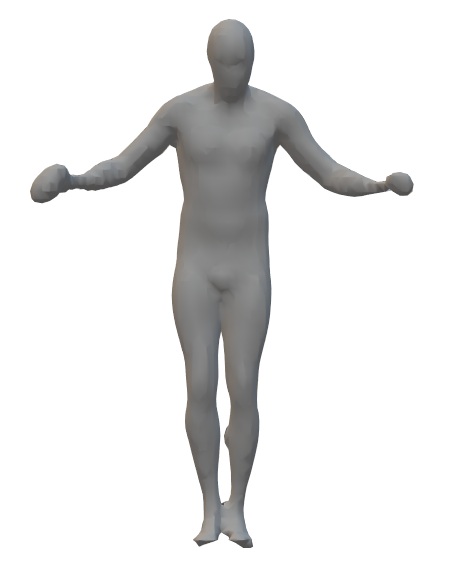}&\includegraphics[ height= 0.7 in]{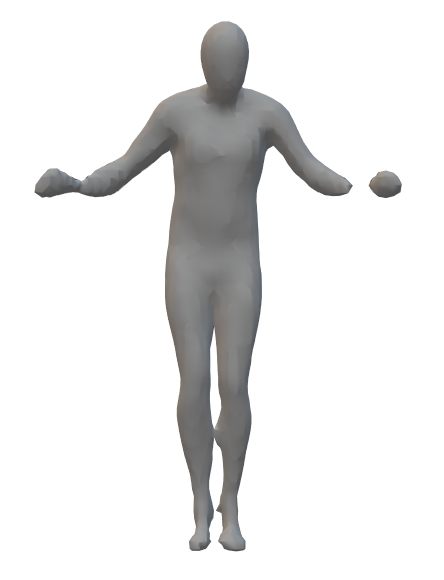}&\includegraphics[ height= 0.7 in]{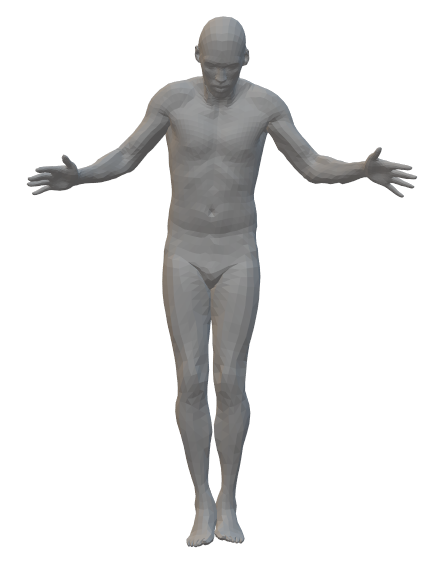}\\
\end{tabular}
\end{adjustbox}
\end{center}
   \caption{Failure cases from the D-Faust dataset. Each row contains three cases with four results per case: (from left to right) input test scan, 2000 epoch baseline SAL reconstruction, 500 epoch LightSAL reconstruction, ground truth. The lowermost-middle case reflects a scenario, where the input scan includes an artifact -- for LightSAL the artifact causes somewhat distortion, whereas for baseline SAL the distortion is significant.}
\label{reconstructed_huma_shape:failure_Cases}
\end{figure*}

\subsection{Notes}
Similar to what was reported \cite{Atzmon_2020_CVPR} by the authors of the baseline SAL architecture, also the proposed architecture has difficulties in reconstructing thin structures. Some failed reconstruction examples are shown in Figure~\ref{reconstructed_huma_shape:failure_Cases} for the D-Faust dataset.

The pretrained SAL model that was used in the experiment for human shape reconstruction (Section~\ref{ssec:firstexperiment}) had been trained by the SAL \cite{Atzmon_2020_CVPR} authors with a batch size of 64, whereas due to GPU memory restrictions the batch size of LightSAL training was restricted to 16. The difference in batch size might have a minor effect to the results. Similarly, due to limited GPU memory, also the reconstruction resolution was limited to 100, instead of 512 that was used in the code by baseline SAL authors.

\section{Discussion}\label{discussion}

Inspecting the proposed LightSAL architecture, one might wonder for which reason the small proposed model performs as well as the baseline SAL model that has a four-fold number of trainable parameters. One explanation for this is that although a large, fully-connected model has the possibility to memorize the training data extensively, this may cause difficulty for the trained model to generalize on unseen test data. Although this observation is empirical, such behavior can be seen from Tables \ref{chamfer_distance:Dfaust_Dataset_unseenHuman}, \ref{chamfer_distance:Dfaust_Dataset_unseenPose}  and \ref{ablation_study:Chamfer_distances}.

Another detail worth discussing is related to the choice of layer types. Here, the recently proposed concept of Convolutional Occupancy Networks \cite{Peng2020ECCV} and PointNet \cite{qi2017pointnet} inspired our work towards adopting a shared MLP architecture using convolutional layers with kernel size 1 instead of a shared MLP constructed with a fully-connected layer.However, for keeping the network size small in terms of parameter count, we selected the 1D convolutional layer type over costlier 2D/3D convolutions \cite{qi2017pointnet,Peng2020ECCV} used in PointNet (a shared MLP), and Convolutional Occupancy network, respectively. Empirically, it was noticed that 1D convolutions (in the encoder case) with kernel size 1$\times$1 perform equally well as a 3$\times$3 kernel, but reduce the number of parameters by a large margin.

Furthermore, the encoder in our work consumes a point cloud that is an unordered and permutation invariant data type. Therefore, a network that is fed point cloud should be able to deal with this unique nature. For this reason PointNet \cite{qi2017pointnet} proposed to use shared MLP followed by a symmetric function (Max Pool). The shared MLP in their work is not a fully connected layer; rather, it is a 2D convolutional layer with kernel size 1. Shared MLP constructed using fully connected layers have been used in the baseline SAL architecture \cite{Atzmon_2020_CVPR}. 
However, based on our experiments, we have empirically shown that permutation and order invariant data can be handled with permutation non-invariant network architectures ( when symmetric function considers two neighbors). We also observe that shared MLP formation with 1x1 convolutional layer is beneficial compared to a fully connected layer for the decoder, which was empirically validated in the appendix section. 
Therefore, a shared MLP constructed using 1x1 convolutional layer with kernel size 1 has potential to be better suited to an implicit decoder than a shared MLP constructed using a fully connected layer. In this work, we have also demonstrated this with empirical evidence.

As a final note, very recently, another novel approach, unsigned distance, \cite{chibane2020ndf} for surface representation has been proposed in the 3D modeling literature, and appears to have promising characteristics for modeling open surfaces and scenes. As future work, our intention is to explore the possibility for adapting the LightSAL architecture to unsigned distance fields.

\section{Conclusion}
In this paper we have presented LightSAL, a lightweight encoder-decoder architecture for implicit reconstruction of 3D shapes, building on the concept of Sign Agnostic Learning. The LightSAL architecture has 75\% less trainable parameters than the baseline SAL architecture, trains 40\% faster per epoch, and provides equivalent reconstruction quality with 500 epochs when compared to the 2000-epoch trained baseline SAL model. The D-Faust dataset with 41k human scans was used for experimental evaluation.

In terms of observed visual quality, baseline SAL occasionally suffers from behavior where the reconstruction converges towards a different shape than what was indicated by the input data. Such unwanted behavior was not observed with LightSAL. In general, LightSAL outperforms baseline SAL clearly in the quality of test-time reconstruction of unseen shapes.

As broader impact of this work we see that LightSAL highlights the importance of studying compact architectures for implicit reconstruction of 3D shapes. LightSAL has demonstrated that even a significant reduction in architecture size can be performed without affecting reconstruction quality. Continuing with this research direction could open up new application areas for implicit shape representation.


\section{Acknowledgment}

This work was partially funded by the Academy of Finland project 334755 CoEfNet. The authors wish to acknowledge CSC – IT Center for Science, Finland, for computational resources, and Antti Kinnunen for technical assistance.

{\small
\bibliographystyle{ieee_fullname}

}

\end{document}